\documentclass{article}

\usepackage{microtype}
\usepackage{graphicx}
\usepackage{subfig}
\usepackage{booktabs} 

\usepackage{hyperref}

\usepackage[accepted]{icml2023}
\usepackage{amsmath}
\usepackage{amssymb}
\usepackage{mathtools}
\usepackage{amsthm}

\usepackage[capitalize,noabbrev]{cleveref}

\usepackage{amsopn,amsmath,amsthm,amssymb}
\usepackage{wrapfig}
\usepackage{multirow}
 \usepackage{mathtools}

\usepackage{anyfontsize}

\usepackage{color}
\usepackage{tikz}
\usetikzlibrary{arrows,shapes,snakes,automata,backgrounds,fit,petri}
\usepackage{adjustbox}

\newcommand{\RN}[1]{%
	\textup{\lowercase\expandafter{\it \romannumeral#1}}%
}

\makeatletter
\newcommand{\distas}[1]{\mathbin{\overset{#1}{\kern\z@\sim}}}%

\usepackage{enumitem}

\usepackage{algorithm}
\usepackage{algorithmic}

\newcommand{\beq}{\vspace{0mm}\begin{equation}}
\newcommand{\eeq}{\vspace{0mm}\end{equation}}
\newcommand{\beqs}{\vspace{0mm}\begin{eqnarray}}
\newcommand{\eeqs}{\vspace{0mm}\end{eqnarray}}
\newcommand{\barr}{\begin{array}}
\newcommand{\earr}{\end{array}}

\newcommand{\R}{\mathbb{R}}

\newtheorem{theorem}{Theorem} 
\newtheorem{lemma}{Lemma}

\ifx\assumption\undefined
\newtheorem{assumption}{Assumption}
\fi

\ifx\definition\undefined
\newtheorem{definition}{Definition}
\fi

\ifx\remark\undefined
\newtheorem{remark}{Remark}
\fi
\newcommand{\norm}[1]{\left\| #1 \right\|}

\newcommand{\Abs}[1]{\left| #1 \right| }

\usepackage{arydshln}

\usepackage{mathtools}

\definecolor{DarkGreen}{rgb}{0.1,0.5,0.1}

\newcommand{\eps}{\epsilon}

\icmltitlerunning{DP-Fast MH}

\begin{document}

\twocolumn[
\icmltitle{DP-Fast MH: Private, Fast, and Accurate Metropolis-Hastings\\ for Large-Scale Bayesian Inference}



\icmlsetsymbol{equal}{*}

\begin{icmlauthorlist}
\icmlauthor{Wanrong Zhang}{equal,harvard}
\icmlauthor{Ruqi Zhang}{equal,purdue}
\end{icmlauthorlist}

\icmlaffiliation{harvard}{Harvard John A. Paulson School of Engineering and Applied Sciences
}
\icmlaffiliation{purdue}{Department of Computer Science, Purdue University
}
\icmlcorrespondingauthor{Wanrong Zhang}{wanrongzhang@fas.harvard.edu}
\icmlcorrespondingauthor{Ruqi Zhang}{ruqiz@purdue.edu}

\icmlkeywords{Machine Learning, ICML}

\vskip 0.3in
]



\printAffiliationsAndNotice{\icmlEqualContribution} 

\begin{abstract}
Bayesian inference provides a principled framework for learning from complex data and reasoning under uncertainty. It has been widely applied in machine learning tasks such as medical diagnosis, drug design, and policymaking.
In these common applications, data can be highly sensitive.
Differential privacy (DP) offers data analysis tools with powerful worst-case privacy guarantees and has been developed as the leading approach in privacy-preserving data analysis. In this paper,
we study Metropolis-Hastings (MH), one of the most fundamental MCMC methods, for large-scale Bayesian inference under differential privacy.
While most existing private MCMC algorithms sacrifice accuracy and efficiency to obtain privacy, we provide the first exact and fast DP MH algorithm, using only a minibatch of data in most iterations. We further reveal, for the first time, a three-way trade-off among privacy, scalability (i.e. the batch size), and efficiency (i.e. the convergence rate), theoretically characterizing how privacy affects the utility and computational cost in Bayesian inference. We empirically demonstrate the effectiveness and efficiency of our algorithm in various experiments.
\end{abstract}

\section{Introduction}

Bayesian inference is one of the most powerful tools in data analytics
and has been applied on a wide variety of data tasks~\citep{gelman1995bayesian}. In practice, computing the posterior
is often intractable and thus requires approximation methods, such as Markov chain Monte
Carlo (MCMC). The Metropolis-Hastings (MH) algorithm~\citep{metropolis1953equation,hastings1970monte} is one of the most commonly used MCMC methods, where it works by conducting an accept/reject step to ensure the target distribution as its equilibrium distribution. By its construction, MH is guaranteed to converge asymptotically to the true posterior.

In many applications of Bayesian inference, privacy becomes a crucial consideration as the data can be highly sensitive. For example, doctors may use Bayesian inference to analyze past patient data and find a pattern to make accurate predictions in medical diagnosis; tech companies may use Bayesian methods to classify spam emails by looking at features in customers' emails. In these tasks, we need new Bayesian inference methods that can give useful inference results under the privacy constraint.

{\em Differential privacy} (DP) has emerged as the gold standard for privacy-preserving data analytics. It is a statistical notion of privacy, which ensures that the output distribution of a computation is robust when we arbitrarily change one person's data. DP algorithms hide individuals' data by introducing randomness. While there exists a few DP MCMC algorithms, they have significant limitations on their utility and applicability. In particular, most work has no guarantees on the asymptotic convergence or the convergence speed, thus could diverge far away from the target posterior in practice~\citep{heikkila2019differentially,li2019connecting,zhang2021differentially}. Some work significantly scarifies the convergence rate for privacy and can not scale to large datasets~\citep{yildirim2019exact}. Moreover, most existing methods operate on the belief that there are trade-offs among privacy, efficiency (i.e. the convergence rate), and scalability (i.e. the batch size) in private Bayesian inference, but none of them formally characterize these trade-offs.

In this work, we solve all the above issues by providing a simple yet powerful DP MCMC algorithm, \emph{DP-Fast MH}, with theoretically guarantees on privacy, utility, and scalability. To the best of our knowledge, our method is
the \emph{first} exact minibatch DP MCMC algorithm, which enables private and accurate Bayesian inference on large-scale datasets. Built upon TunaMH ~\citep{zhang2020asymptotically} which is a non-private exact minibatch MH method, we developed DP-Fast MH by injecting an appropriate amount of Gaussian noise in the MH algorithm while still keeping the reversibility of the Markov chain. Notably, we show that due to the inherent randomness in the accept/reject step in MH, \emph{privacy for free} can be achieved under certain cases without adding additional noise. This finding drastically improves the privacy-utility trade-off in previous DP MH methods. Theoretically, we give privacy guarantees (Theorem \ref{thm.privacy}) following subsampling analysis and instantiated Gaussian mechanisms. We also show the exactness of our algorithm (Theorem \ref{thm.asymptotic}), which guarantees the asymptotic convergence, meaning that the distribution is close to the target distribution when the number of iterations gets large. Our convergence rate theorem statement (Theorem \ref{thm.convergence-rate}) further quantifies the cost of privacy in terms of the convergence rate and batch size. 
Empirically, we demonstrate the accuracy of our estimated distribution on several common Bayesian inference tasks. 
We summarize our contributions as follows:

\begin{itemize}
    \item We introduce DP-fast MH, the first private, asymptotically-exact, minibatch Metropolis-Hastings algorithm. Taking advantage of the inherent randomness in MH, we achieve privacy for free under certain conditions, which greatly improves the privacy-utility trade-off in existing differentially private MH methods.
    \item We prove $(\epsilon,\delta)$-DP guarantees, asymptotic convergence, and convergence rate bound for DP-fast MH, formally characterizing a three-way trade-off among privacy, scalability (i.e. the batch size),
and efficiency (i.e. the convergence rate).
    \item We empirically validate our theorems, and demonstrate that our algorithm significantly outperforms previous methods in terms of both estimation accuracy and computational cost under various settings.

\end{itemize}

\section{Related Work}

Differentially private posterior sampling has recently gained attractions, with the methods falling into three categories: DP Metropolis-Hastings~\citep{heikkila2019differentially,yildirim2019exact}, DP Gradient-based MCMC~\citep{li2019connecting,zhang2021differentially,raisa2021differentially}, and DP Exact Posterior Sampling~\citep{dimitrakakis2014robust,wang2015privacy,foulds2016theory,zhang2016differential}

\paragraph{DP Metropolis-Hastings}
Differentially private Metropolis-Hastings has been developed in~\citet{heikkila2019differentially,yildirim2019exact}. \citet{heikkila2019differentially} gives an algorithm based on the Barker acceptance test, which can be used with a minibatch of data. However, the algorithm has an asymptotic bias, making it unreliable in practice: the stationary distribution might be arbitrarily far away from the target distribution. Moreover, there is no theoretical convergence rate guarantee and setting the privacy budget upfront is hard.
\citet{yildirim2019exact} considers the penalty algorithm~\citep{ceperley1999penalty} where Gaussian noise is added to the log-likelihood difference. This method converges to the correct distribution asymptotically, but the convergence speed decreases significantly due to adding a large amount of noise. 
Also, this method uses the entire dataset to evaluate the acceptance probability in each iteration, so it is computationally infeasible with large datasets. 

\paragraph{DP Gradient-based MCMC}
Another line of work studies (stochastic) gradient-based MCMC methods~\citep{li2019connecting,zhang2021differentially,raisa2021differentially,deng2021convergence}. Differentially private Stochastic Gradient MCMC (SGMCMC), including Stochastic Gradient Langevin Dynamics (SGLD)~\citep{welling2011bayesian} and other variants, are shown to satisfy differential privacy guarantees by default. However, DP SGMCMC has an asymptotic bias and also lacks non-asymptotic convergence guarantees. It is worth noting that DP SGLD may have unbounded private loss even if the exact posterior is differentially private as desired~\citep{heller2021can}. \citet{raisa2021differentially} develops a DP version of Hamiltonian Monte Carlo~\citep{neal2011mcmc} whose MH step is based on~\citet{yildirim2019exact}, thus inheriting the drawbacks of slow convergence and expensive computation.

\paragraph{DP Exact Posterior Sampling}
\citet{dimitrakakis2014robust,wang2015privacy,foulds2016theory,zhang2016differential} consider differentially private posterior sampling, where the posterior is assumed to be obtained exactly. In turn, they only offer conditional privacy guarantees, which rely on the exact sampling assumption. The posterior is often intractable in practice, and it is usually impossible to validate the assumption in practical algorithms.

\section{Preliminaries}
This section provides necessary background on Metropolis-Hastings and differential privacy for understanding our algorithm and theoretical results.

\subsection{Metropolis-Hastings}
In Bayesian inference, given a dataset $D = \{x_i\}_{i=1}^N$ and a $\theta$-parameterized model, we aim to compute the posterior distribution 
\begin{align*}
    \pi(\theta) &\propto \exp\left(-\sum_{i=1}^N U_i(\theta)\right)\\&=\exp\left(\sum_{i=1}^N  \log p(x_i|\theta) +\log p(\theta)\right)
\end{align*}

where $U_i(\theta)$ is the energy function, $p(x_i|\theta)$ is the likelihood and $p(\theta)$ is the prior.
The Metropolis-Hastings (MH) algorithm approximates the posterior by sampling from it. In each step, MH generates a proposal $\theta'$ from some distribution $q(\cdot\mid\theta)$ and accepts $\theta'$ with probability
\[
a(\theta,\theta')=\min\left(1, \exp\big( \sum_{i=1}^N (U_i(\theta) - U_i(\theta')) \big) \cdot \frac{q(\theta|\theta')}{q(\theta'|\theta)}\right).
\]

The construction of $a(\theta,\theta')$ ensures the reversibility of the Markov chain, which is a common condition to guarantee correct asymptotic convergence.

\paragraph{TunaMH} The acceptance probability $a(\theta,\theta')$ in MH requires computing a sum over the entire dataset, which could be prohibitively expensive on large datasets. A common way to solve this issue is to use a minibatch approximation. In this paper, we build our DP MH algorithm upon a recently developed exact minibatch MH method, TunaMH~\citep{zhang2020asymptotically}, which uses Poisson auxiliary variables to enable minibatching while still keeping the correct asymptotic convergence. The algorithm of TunaMH is outlined in Algorithm~\ref{alg:poisson-mh}. At each iteration, TunaMH forms a minibatch $\mathcal{I}$ by sampling a batch size $B$ from a Poisson distribution, and then sample (with replacement) a minibatch of size $B$ to compute the acceptance probability. The expected batch size has been shown to be upper bounded by $\mathbf{E}[B] = \lambda + C M(\theta, \theta')$. Different from the original TunaMH, we keep the hyperparameter $\lambda$ to control the batch size rather than setting $\lambda=\chi C^2M^2(\theta,\theta')$ as suggested in the paper. The reason of keeping $\lambda$ independent of $CM(\theta,\theta')$ is to bound the sensitivity of energy difference, which will be seen later in Section~\ref{sec:dpfast-mh}.

\begin{algorithm}[t]
  \caption{TunaMH}
  \begin{algorithmic}
    \label{alg:poisson-mh}
    \STATE \textbf{given:} initial state $\theta \in \Theta$; proposal dist. $q$; hyperparameter $\lambda$; Asm.~\ref{assump} parameters $c_i$, $C$, $M$
    \LOOP
      \STATE \textbf{propose} $\theta'\sim q(\cdot|\theta)$ and \textbf{compute} $M(\theta, \theta')$
      \STATE \textbf{sample} $B \sim \text{Poisson}\left( \lambda + CM(\theta,\theta')\right)$
      \STATE  \textbf{obtain} batch indices $\mathcal{I} \leftarrow \texttt{Batch}(B)$
      
      \STATE \textbf{obtain} MH Ratio $r\leftarrow \texttt{MHRatio}(\mathcal{I},0)$
      
      \STATE \textbf{with probability} $\min(1,r)$, set $\theta \leftarrow \theta'$
    \ENDLOOP
    \vspace{0.5em}
    \STATE Function \texttt{Batch}($B$)
      \STATE \textbf{initialize minibatch indices} $\mathcal{I} \leftarrow \emptyset$
      \FOR{$b \in \{1,\ldots,B\}$}
        \STATE \textbf{sample} $i_b$ such that $\mathbf{P}(i_b = i) = c_i/C$, for $i=1\ldots N$
        \STATE \textbf{with probability} $\frac{\lambda c_{i_b}  + \frac{C}{2}(U_{i_b}(\theta') - U_{i_b}(\theta) + c_{i_b}M(\theta,\theta'))}{\lambda c_{i_b} + c_{i_b} CM(\theta, \theta')}$ \textbf{add} $i_b$ to $\mathcal{I}$  
      \ENDFOR
      \STATE return $\mathcal{I}$
      
      \vspace{0.5em}
    \STATE Function $\texttt{MHRatio}(\mathcal{I},\tau)$
    \STATE {\small{$r \leftarrow
        \exp\left(2 \sum_{i \in \mathcal{I}} \operatorname{artanh}\left(
        \frac{C \left(U_i(\theta) - U_i(\theta') \right)}{c_i (2 \lambda + C M(\theta,\theta'))} 
    \right) +\tau \right)
        \cdot \frac{q(\theta|\theta')}{q(\theta'|\theta)}$}
        \STATE return $r$}
  \end{algorithmic}
\end{algorithm}

\begin{algorithm}[t]
  \caption{DP-Fast MH}
  \begin{algorithmic}
    \label{alg:dp-fastmh}
    \STATE \textbf{given:} initial state $\theta \in \Theta$; proposal dist. $q$; hyperparameter $\lambda$; Asm.~\ref{assump} parameters $c_i$, $C$, $M$, {\color{blue}privacy parameters $\eps,\delta$, batch size upper bound $K$}.
    \STATE {\color{blue}Set noise scale $\sigma_2=\sqrt{2\log\frac{1.25}{\delta}}/\eps$ and $\sigma_1=6K\max_i c_i\sqrt{2\log\frac{2.5 K\max_i c_i}{\delta C}}/(\eps C)$}.
    \LOOP
      \STATE \textbf{propose} $\theta'\sim q(\cdot|\theta)$ and \textbf{compute} $M(\theta, \theta')$
      \STATE \textbf{sample} $B \sim \text{Poisson}\left( \lambda  + CM(\theta,\theta')\right)$
      \IF[{{\color{red} Run minibatch MH}}]{{\color{blue}$B<K$}} 
      \STATE  \textbf{obtain} batch indices $\mathcal{I} \leftarrow \texttt{Batch}(B)$
      
    {\color{blue}\IF[{{\color{red} Privacy for free}}]{$\Delta(\ell_1)\le \eps C/(6 K\max_i c_i)$}
      \STATE Let $\xi=0$.
      \ELSE
      \STATE Sample $\xi\sim \mathcal{N}(0, \sigma_1^2 \Delta(\ell_1)^2)$.
      \ENDIF}
      
      \STATE \textbf{obtain} MH Ratio $r\leftarrow \texttt{MHRatio}(\mathcal{I},\xi-\frac{\sigma_1^2 \Delta(\ell_1)^2}{2})$
     \ELSE[{{\color{red} Run full-batch MH}}]
     {\color{blue}
     \IF[{{\color{red} Privacy for free}}]{$\Delta(\ell_2)\le \eps$}
     \STATE Let $\xi=0$.
     \ELSE
     \STATE Sample $\xi\sim \mathcal{N}(0,\sigma_2^2 \Delta(\ell_2)^2)$
     \ENDIF
     \STATE $r \leftarrow
        \exp\big( \sum_{i=1}^N (U_i(\theta) - U_i(\theta'))+ \xi -\frac{\sigma_2^2 \Delta(\ell_2)^2}{2} \big) \cdot \frac{q(\theta|\theta')}{q(\theta'|\theta)}$
        }
     \ENDIF

      \STATE \textbf{with probability} $\min(1,r)$, set $\theta \leftarrow \theta'$
      
    \ENDLOOP
  \end{algorithmic}
\end{algorithm}

\subsection{Differential Privacy}

Differential privacy is a mathematical framework for privacy-preserving data analysis. It
bounds the maximum amount that one person's data can affect the analysis performed on the dataset. We say two datasets $D, D'$ are neighboring if they arbitrarily differ in the values at most one entry. 

\begin{definition}[Differential Privacy \cite{DMNS06}]\label{def.dp}
	A randomized algorithm $\mathcal{M}: \mathcal{D} \rightarrow \mathcal{R}$ is \emph{$(\epsilon,\delta)$-differentially private} if for every pair of neighboring datasets $D,D' \in \mathcal{D}$, and for every subset of possible outputs $\mathcal{S} \subseteq \mathcal{R}$,
	$\Pr[\mathcal{M}(D) \in \mathcal{S}] \leq \exp(\epsilon)\Pr[\mathcal{M}(D') \in \mathcal{S}] + \delta.$
\end{definition}

Differentially private mechanisms typically add noise that scales with the sensitivity of the function being evaluated and the idea is to use randomness to hide each individual's contribution. The sensitivity of a function $f$ is defined as the maximum change in $f$ between two neighboring sets: $\Delta f = \max_{D,D' \text{ neighbors}} | f(D) - f(D')|$
The Gaussian mechanism with parameters $(\eps, \delta)$ takes in a function $f$, dataset $D$, and outputs $f(X)+\mathcal{N}(0,\sigma^2)$, where $\sigma=\sqrt{2\log(1.25/\delta)}\Delta f/\eps$. 

\begin{theorem}[Privacy of Gaussian Mechanism \cite{dwork2014algorithmic}]\label{thm.dpgaussian}
	For any $\eps, \delta \in (0,1)$, the Gaussian Mechanism with parameter $\sigma= \sqrt{2\log(1.25/\delta)}\Delta f/\eps$ is $(\eps,\delta)$-differentially private.
\end{theorem}

Differential privacy is also robust under post-processing, which means once an output is differentially private, we can perform more computations based on the private output without leaking any more information than the privacy parameters. This property allows us to build our DP MH algorithm that privatizes the intermediate samples.

Differentially private algorithms compose adaptively. The advanced composition \cite{dwork2010boosting} shows that the privacy parameter is $(O(\eps\sqrt{T\log 1/\delta}), T\delta)$ under $T$ iterations. Recently, several relaxations of differential privacy including R\`enyi differential privacy (RDP) \cite{mironov2017renyi}, and $f$-differential privacy ($f$-DP) \cite{dong2019gaussian} allow for tighter reasoning about composition. Moreover, reasoning the overall privacy guarantee of our entire algorithm is similar to that of the DP stochastic gradient descent (DP-SGD) algorithm~\cite{abadi2016deep}. The moments accountant technique in that paper was proposed specifically for analyzing DP-SGD, which gives a tighter bound: $(O(\eps\sqrt{T}), \delta)$ under $T$ iterations.

\section{DP-Fast MH: Private, Fast, and Accurate Metropolis-Hastings}\label{sec:dpfast-mh}

In this section, we present our private exact minibatch Metropolis-Hastings algorithm, termed \emph{DP-Fast MH}.
Following previous work on exact minibatch MH methods~\citep{cornish2019scalable, zhang2020asymptotically}, we have the following assumption on the posterior distribution.

\begin{assumption}\label{assump}
For some constants $c_1, \ldots, c_N \in \R_+$, with $\sum_i c_i = C$, and symmetric function $M: \Theta \times \Theta \rightarrow \R_+$, for any $\theta, \theta' \in \Theta$, the energy difference is bounded by
$|U_i(\theta) - U_i(\theta')|\le c_i M(\theta,\theta')$.
\end{assumption}

This assumption holds for many common Bayesian inference problems. For example, if each energy function $U_i$ is $L_i$-Lipschitz continuous, then we can set $c_i = L_i$ and $M(\theta, \theta') = \|\theta - \theta'\|$. 

Assumption \ref{assump} assumes a local bound for the energy function, however, to ensure differential privacy, we need a global bound. Hence, we introduce Assumption \ref{assump2}, which has been used in all previous DP MH work~\citep{heikkila2019differentially,yildirim2019exact}. 
\begin{assumption}\label{assump2}
We additionally assume $M(\theta,\theta')\le A, \forall \theta, \theta' \in \Theta$ for some constant $A$. 
\end{assumption}

Our algorithm is given formally in Algorithm \ref{alg:dp-fastmh} where the differences compared to the non-private baseline TunaMH are highlighted in blue color. In particular, there are three key differences to privatize MH while maintaining its original utility, which we discuss in details separately below. 

\paragraph{Upper Bound on the Batch Size} First, we add an additional hyperparameter $K$ to decide whether to use a minibatch or the entire dataset. 
If the batch size $B$ is less than the upper bound $K$, the algorithm will form a minibatch as in TunaMH; otherwise, it uses the entire dataset as in the standard MH. This modification is critical to guarantee privacy because the batch size $B$ is a random variable ranging from $0$ to $\infty$. Since we form the minibatch by sampling with replacement, it has two competing effects on privacy. On one hand, the possibility that an individual is sampled multiple times incurs additional privacy loss. On the other hand, the possibility that an individual is not sampled into the minibatch amplifies privacy. Intuitively, if the batch size is small, then the second effect dominates. Therefore, we limit the upper bound of the batch size to be a constant $K$. 

In practice, $K$ is often large enough, and most iterations will use minibatches. One may wonder whether we can use sampling without replacement to avoid the first effect. As shown in~\citet{zhang2020asymptotically}, doing this will make the Markov chain not satisfy the reversibility, thus the algorithm will not preserve the correct stationary distribution.

\paragraph{Added Gaussian Noise} Second, DP-fast MH instantiates the Gaussian mechanism to preserve privacy, which adds Gaussian noise to the energy differences $\ell_1=2 \sum_{i \in \mathcal{I}} \operatorname{artanh}\left(
        \frac{C \left(U_i(\theta) - U_i(\theta') \right)}{c_i (2 \lambda + C M(\theta,\theta'))} 
    \right)$ (or $\ell_2=\sum_{i=1}^N (U_i(\theta) - U_i(\theta'))$ when using the standard MH).   
We note that similar idea of adding Gaussian noise appeared in the penalty algorithm \citet{ceperley1999penalty} in the non-private literature and was used as a DP MH algorithm later in~\citet{yildirim2019exact}. The Gaussian noise is scaled with the sensitivity of the energy difference, which is given as follows.

\begin{lemma}\label{lemma.sen}
The sensitivity of the energy difference is
\begin{align}
\Delta(\ell_1)&=2\log \left(1+ \frac{CM(\theta,\theta')}{\lambda }\right)\\
\Delta(\ell_2)&=2\max_i c_i M(\theta,\theta').
\end{align}
\end{lemma}

Both $\Delta (\ell_1)$ and $\Delta (\ell_2)$ monotonically increase as $M(\theta, \theta')$ and $C$ (or $\max_i c_i)$ increase. 
This indicates that the added noise per iteration is large when the proposed $\theta'$ is far away from the current $\theta$ or the Lipschitz constant of the energy function is large. Another interesting observation is that $\Delta(\ell_2)$ is generally smaller than $\Delta(\ell_1)$. This indicates that, to make one iteration the same level of privacy, we need to add \emph{more} noise in the minibatch MH than in the standard MH and more noise means slower convergence. While using minibatches causes a decrease in the convergence rate of MCMC in the non-private setting~\citep{zhang2020asymptotically}, here, we show that the privacy constraint further amplifies this decrease. To derive noise scale in Algorithm \ref{alg:dp-fastmh}, we further show that the sampling with replacement step affects the privacy by a $6K\max_i c_i/C$ factor (see the proof of Theorem \ref{thm.privacy} in Appendix \ref{app.privacyproof} for details). 
Thus, by setting the standard deviation of the Gaussian mechanism to $\eps C/(6 K\max_i c_i)$ when using minibatches, we maintain the same level of privacy per iteration at $\eps$ as using full-batch data.

After obtaining a Gaussian noise $\xi$, we add it to the energy difference and additionally subtract a noise variance$/2$. The subtraction is to ensure that the private Markov chain still satisfies the reversibility and thus is guaranteed to be close to the target distribution along running~\cite{ceperley1999penalty,yildirim2019exact}. 

\paragraph{Privacy for Free} Third, we \emph{only} add the additional Gaussian noise when the sensitities are larger than some constants. This is because we can achieve privacy for free due to the \emph{intrinsic} randomness in the accept/reject step of the MH methods. Specifically, in the MH correction step, we 
\texttt{accept} or \texttt{reject} the proposed $\theta'$ with a probability depends on the energy functions $U_i$.
For any two neighboring datasets $D$ and $D'$, we consider the following probability ratio
\begin{align*}
\frac{\Pr(\texttt{accept }\theta'|D)}{\Pr(\texttt{accept }\theta'|D')}
=\frac{\min(0,r(D)}{\min(0,r(D') }
\le  \exp(\Delta(\ell_1)).
\end{align*}
Therefore, if $\Delta(\ell_1)\le \eps C/(6 K\max_i c_i)$, or similarly, $\Delta(\ell_2)\le \eps $, then the inherent randomness in the MH correction step can achieve differential privacy guarantee, and no additional noise is needed!
Since $\Delta(\ell_1)$ and $\Delta(\ell_2)$ depend on the proposed $\theta'$, we compare them against the privacy parameter at each iteration.

As a comparison, \citet{yildirim2019exact} does not fully leverage the inherent privacy guarantees in the MH algorithm. Therefore, they added unnecessary noise given a target privacy level, which significantly slows down the convergence of the Markov chain as shown in Section~\ref{sec:exp}. The Barker method \citet{heikkila2019differentially} relies on the element of randomness in the accept-reject decision, while the level of privacy is directly tied to the quantity of available randomness.
Our method does not suffer from this limitation as we can add more Gaussian noise for stronger privacy without sacrificing convergence to the target.

In summary, at each iteration, DP-fast MH first samples a minibatch size and checks if it uses a minibatch of data or full-batch data. Then it checks whether to require additional Gaussian noise. If so, it will instantiate the Gaussian mechanism which adds Gaussian noise to the energy difference function. Finally, it chooses \texttt{accept} or \texttt{reject} $\theta'$ based on the noisy acceptance probability. Compared to TunaMH, our algorithm only introduces one additional hyperparameter $K$, which upper bounds the minibatch size.

\section{Theoretical Analysis on Privacy and Convergence}

\subsection{Privacy Analysis}

As a starting point, when the sensitivity of the energy differences $\Delta(\ell_1)$ and $\Delta(\ell_2)$ are small, we get privacy for free from the randomness of accepting the proposed $\theta'$. It suffices to analyze privacy that comes from the 
Gaussian mechanism, and privacy amplification by the sampled minibatch for each iteration. While the subroutine is similar to DP-SGD~\cite{abadi2016deep}, our privacy guarantee requires new analysis, since the batch size in our algorithm is a random variable (in contrast to a constant), the minibatch is formed by sampling with replacement (in contrast to without replacement), and different samples have different probabilities being chosen ((in contrast to uniform sampling). Finally, We also account for the possibility of using the standard MH, where the privacy is ensured by instantiating the Gaussian mechanism again. 
We provide the privacy guarantee for each iteration in Theorem \ref{thm.privacy}, and the formal proof is deferred to Appendix \ref{app.privacyproof}. 

\begin{theorem}[Privacy]\label{thm.privacy}
Under Assumption~\ref{assump} and \ref{assump2}, DP-fast MH (Algorithm \ref{alg:dp-fastmh})
is $(\eps,\delta)$-differentially private per iteration.
\end{theorem}

We remark that one can obtain the total privacy loss by any composition theorems over $T$ iterations. Since these bounds are usually loose and the existing DP MH algorithms~\citep{yildirim2019exact, heikkila2019differentially} use different approaches to account for the total privacy loss, we compare our DP-Fast MH algorithm against theirs by fixing the privacy budget for each iteration and the total number of iterations.

\subsection{Convergence Analysis}
Unlike previous DP MCMC work which typically do not come with convergence analysis, we provide both asymptotic and non-asymptotic convergence guarantees for our method. To analyze asymptotic convergence, we first notice that
the way we add a Gaussian noise inside the exponential function does not affect the reversibility of the Markov chain~\citep{ceperley1999penalty}. Then, we show that the combination of the minibatch MH and full-batch MH still results in a reversible transition kernel, since the probability of deciding whether to use minibatches $\Pr(B<K)$ is symmetric in $\theta$ and $\theta'$. Based on the above two reasons, DP-fast MH still converges to the target distribution asymptotically. Theorem~\ref{thm.asymptotic} formally describes our method’s asymptotic accuracy.

\begin{theorem}[Asymptotic Convergence]\label{thm.asymptotic}
Under Assumption~\ref{assump} and ~\ref{assump2}, DP-fast MH (Algorithm \ref{alg:dp-fastmh}) is reversible with the stationary distribution $\pi$.
\end{theorem}

The proof is given in Appendix~\ref{app:asym-convergence}. This theorem guarantees that the posterior approximation from our algorithm  will be closer to the
target distribution when the number of iterations gets larger. 

We now turn our attention to the non-asymptotic convergence of DP-fast MH. Although the correct asymptotic convergence of our method preserves, the convergence rate could be significantly affected by the privacy constraint, which may greatly slow down the algorithm's convergence in practice. Luckily, we show below that the convergence rate of DP-fast MH compared to the standard full-batch non-private MH can only be slowed down by at most a constant factor. We quantify the convergence rate using \emph{spectral gap}, which has been widely used in
the MCMC literature~\citep{hairer2014spectral,levin2017markov,zhang2019poisson,zhang2020asymptotically}. The Markov chain with a larger spectral gap will converge faster.

\begin{theorem}[Convergence Rate]\label{thm.convergence-rate}
Under Assumption~\ref{assump} and ~\ref{assump2}, let $\bar{\gamma}$ denote the
spectral gap of DP-fast MH (Algorithm \ref{alg:poisson-mh}), and let $\gamma$ denote the spectral gap of the standard MH with the same target
distribution and the proposal distribution. Then
\begingroup\makeatletter\def\f@size{9}\check@mathfonts
\def\maketag@@@#1{\hbox{\m@th\large\normalfont#1}}
\begin{align*}
   \frac{\bar{\gamma}}{\gamma} 
   &\ge
  \left(1-\Phi\left(\frac{216K^2\max_i c_i^2\log\frac{2.5 K\max_i c_i}{\delta C}\log^2 \left(1+ \frac{CA}{\lambda}\right)}{\eps^2 C^2}\right)\right) 
  \\&\cdot \exp\left(-\frac{C^2A^2}{\lambda } -2\sqrt{\frac{C^2A^2}{\lambda } \log 2}\right)
   .
\end{align*}\endgroup
\end{theorem}
where $\Phi$ denotes the cumulative distribution function of the standard Gaussian distribution. This theorem shows the relative convergence rate of DP-fast MH with the standard MH.
There are three main takeaways from the theorem. First, this theorem shows that when either the privacy hyperparameter $\eps$ or $\delta$ becomes small, the convergence rate becomes small, characterizing how much the privacy constraint slows down the convergence speed of the Markov chain. 

Furthermore, it reveals a new \emph{three-way} trade-off among privacy, batch size, and convergence rate: given a target privacy, a smaller batch size (i.e. larger $K$ or smaller $\lambda$) results in a worse convergence rate (i.e. the lower bound for $\bar{\gamma}/\gamma$ becomes smaller). This trade-off shows that there exists a sweet spot for the hyperparameters $K$ and $\lambda$ such that the chain achieves an optimal performance considering both estimation accuracy and computational costs. It is generally difficult to obtain theoretical optimal values and we discuss how to tune them in practice in Section~\ref{sec:exp}. 

Finally, our bound shows that when $C$ ( the Lipschitz constant of the energy function) or $A$ (the distance between $\theta$ and $\theta'$) becomes larger, the convergence rate becomes worse, indicating that DP-fast MH is more efficient when the target posterior has a small log-Lipschitz constant or when the proposed $\theta'$ is closer to the current position.

\section{Experiments}\label{sec:exp}

We experimentally demonstrate the accuracy, efficiency and scalability of DP-fast MH under different privacy budget. 
In all experiments, we set the privacy parameter $\delta=10^{-5}$.
In Section~\ref{sec:mog}, we compare DP-fast MH with existing methods under two settings: truncated Gaussian mixture and logistic regression on the MNIST dataset .
In Section \ref{sec:choicek}, we discuss the effect of the newly introduced hyperparameter $K$. For other hyperparameters in the algorithm which are already presented in the baseline method TunaMH, we set the values following \citet{zhang2020asymptotically}. We released the code at \url{https://github.com/ruqizhang/dpfastmh}.
\begin{figure*}
	\vspace{-0mm}\centering
	\begin{tabular}{ccccc}		
		\hspace{-5mm}
		\includegraphics[width=0.25\linewidth]{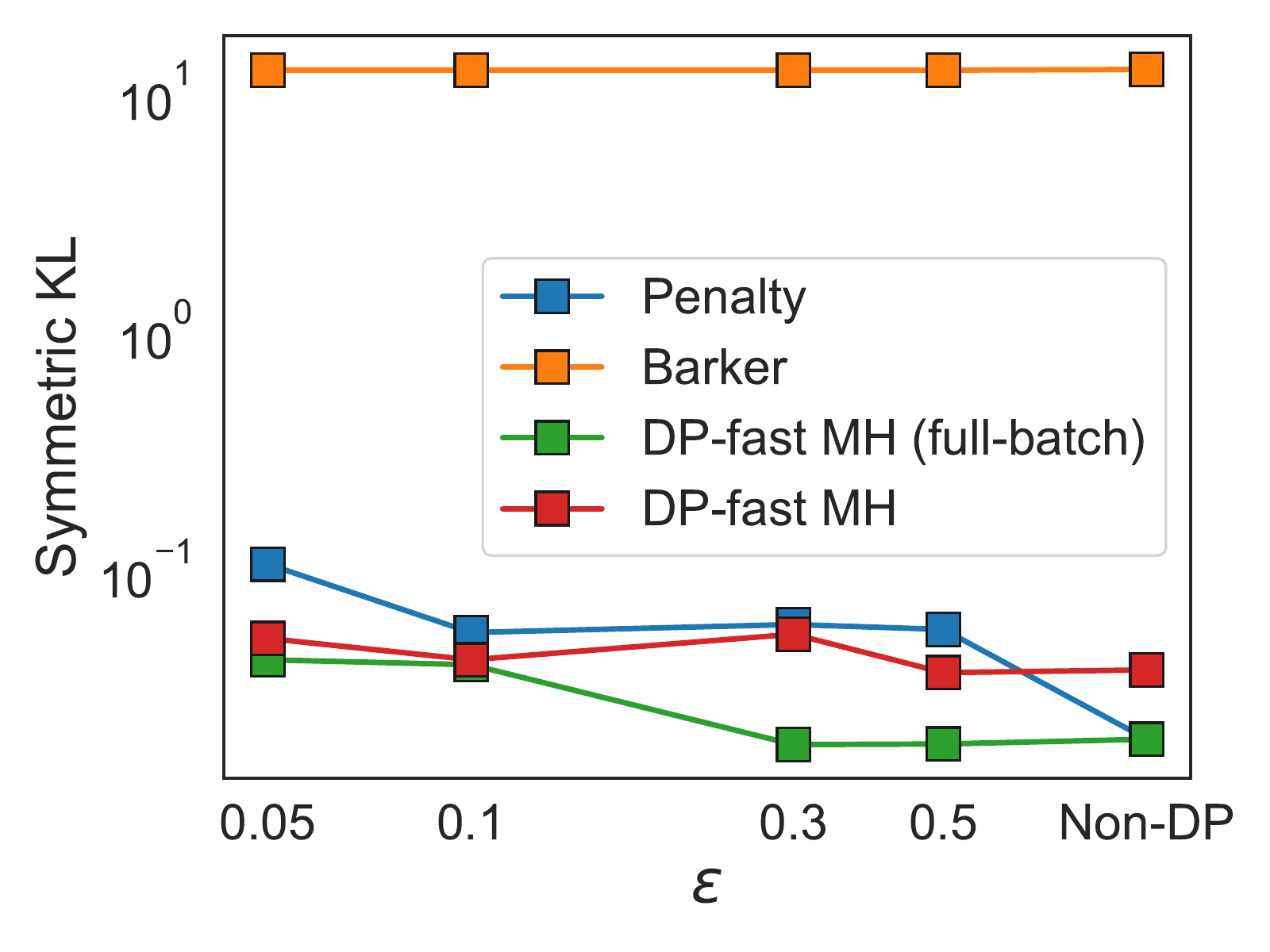}  &
        \hspace{-5mm}
        \includegraphics[width=0.25\linewidth]{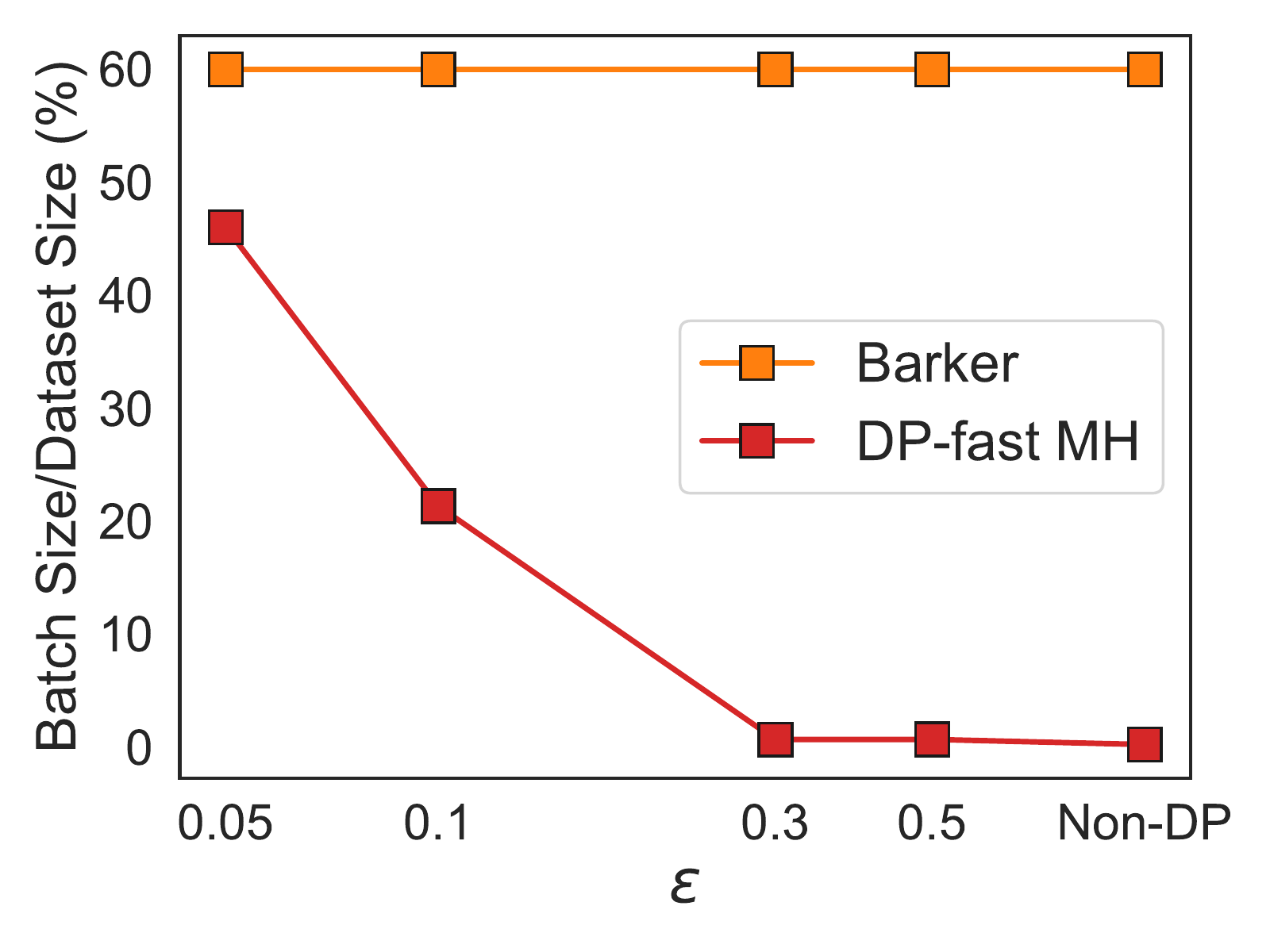}  &
        \hspace{-5mm}
        \includegraphics[width=0.25\linewidth]
        {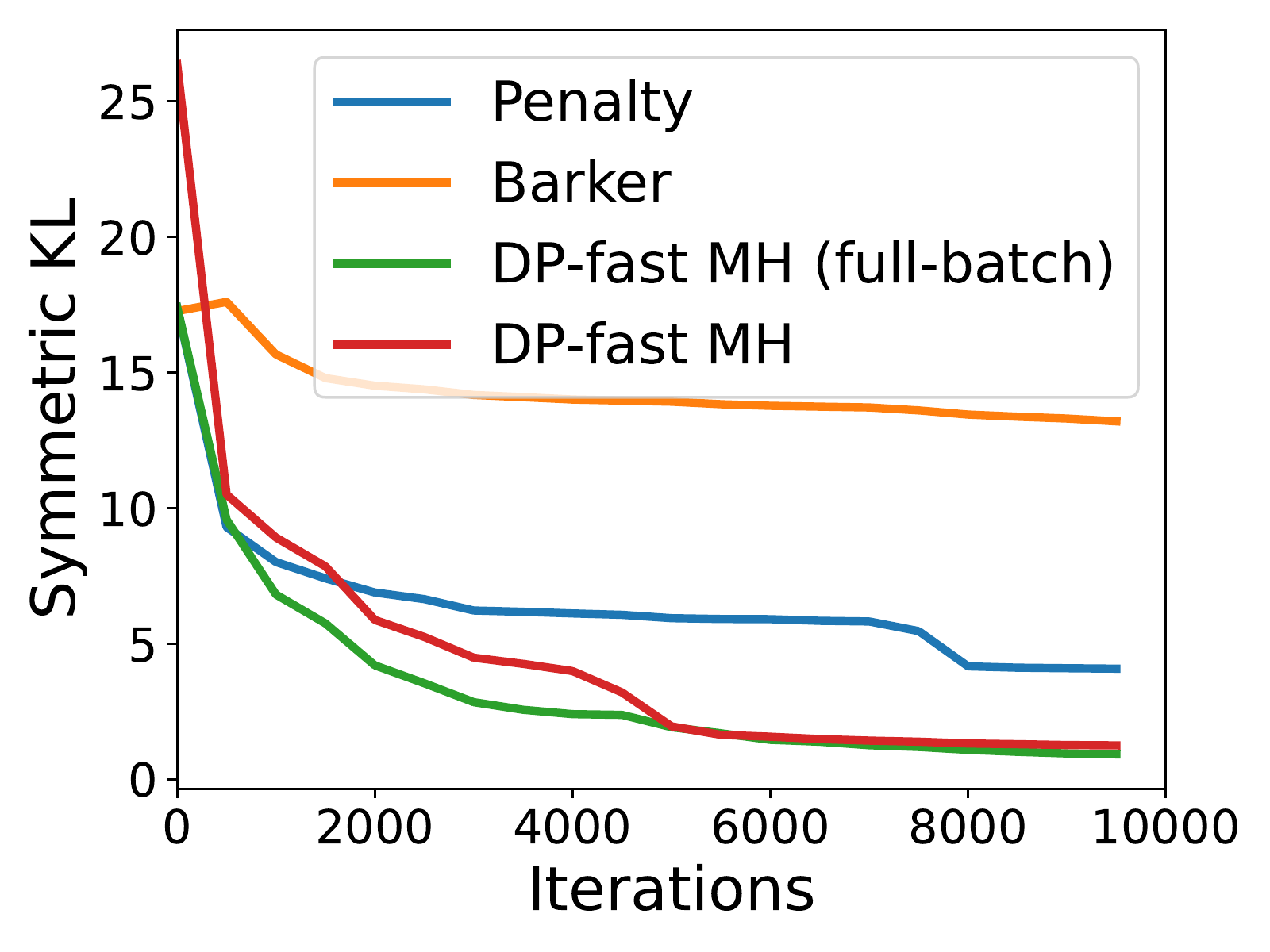} &
        \hspace{-4mm}
        \includegraphics[width=0.25\linewidth]{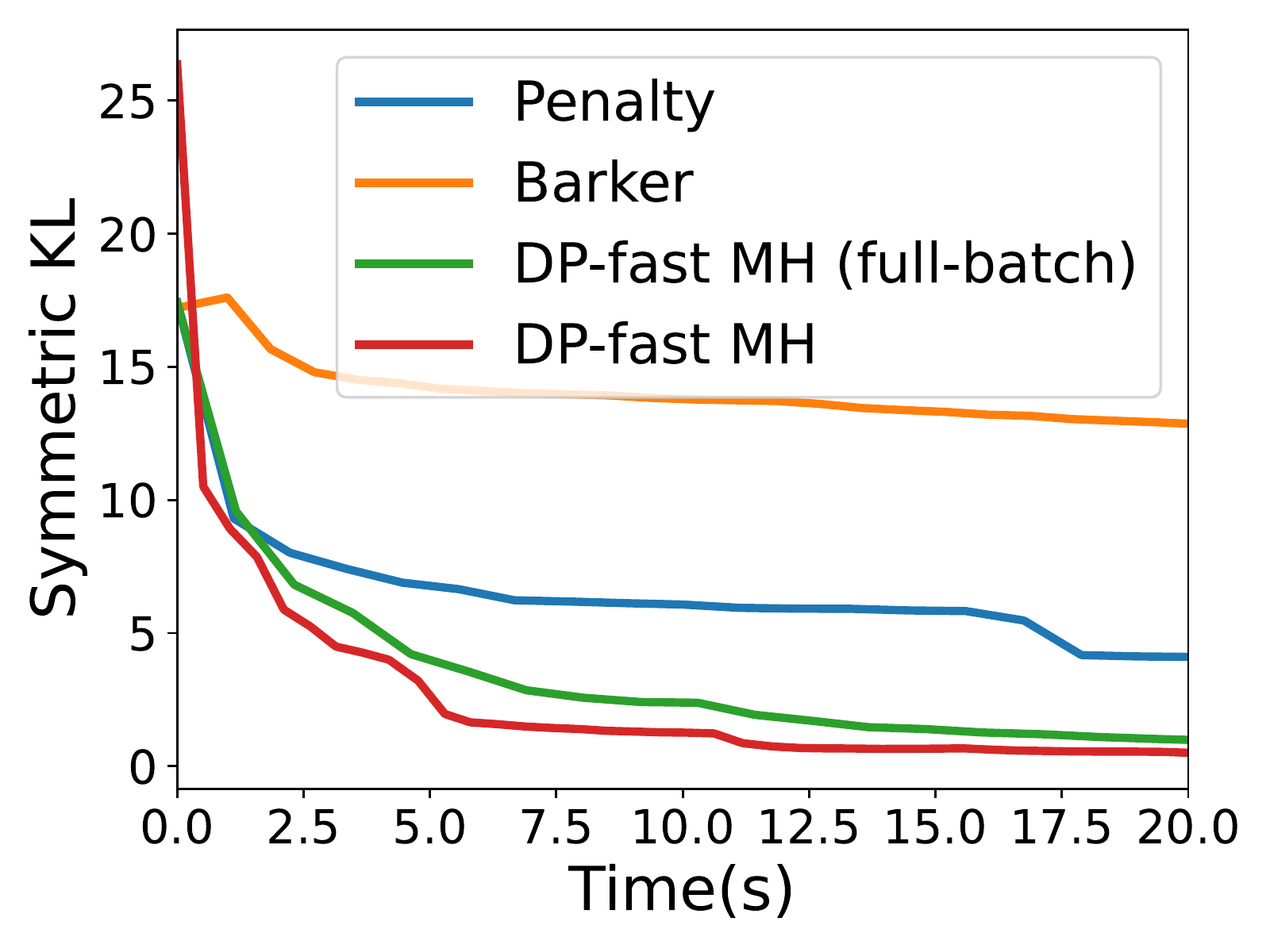}
        \\
		(a)&(b)&(c)&(d) 	\\

	\end{tabular}
	\caption{Truncated Gaussian mixture. (a) Symmetric KL vs. privacy. (b) Batch size vs. privacy. (c)\&(d) When privacy $\epsilon=0.05$, symmetric KL vs. iterations and time. DP-fast MH has the best results in terms of estimation accuracy, scalability, and efficiency. }
	\label{fig:mog}
\end{figure*}

		\begin{figure*}
            \vspace{-5mm}
			\centering
            \subfloat[][Ground Truth]{\includegraphics[width=4.4cm]{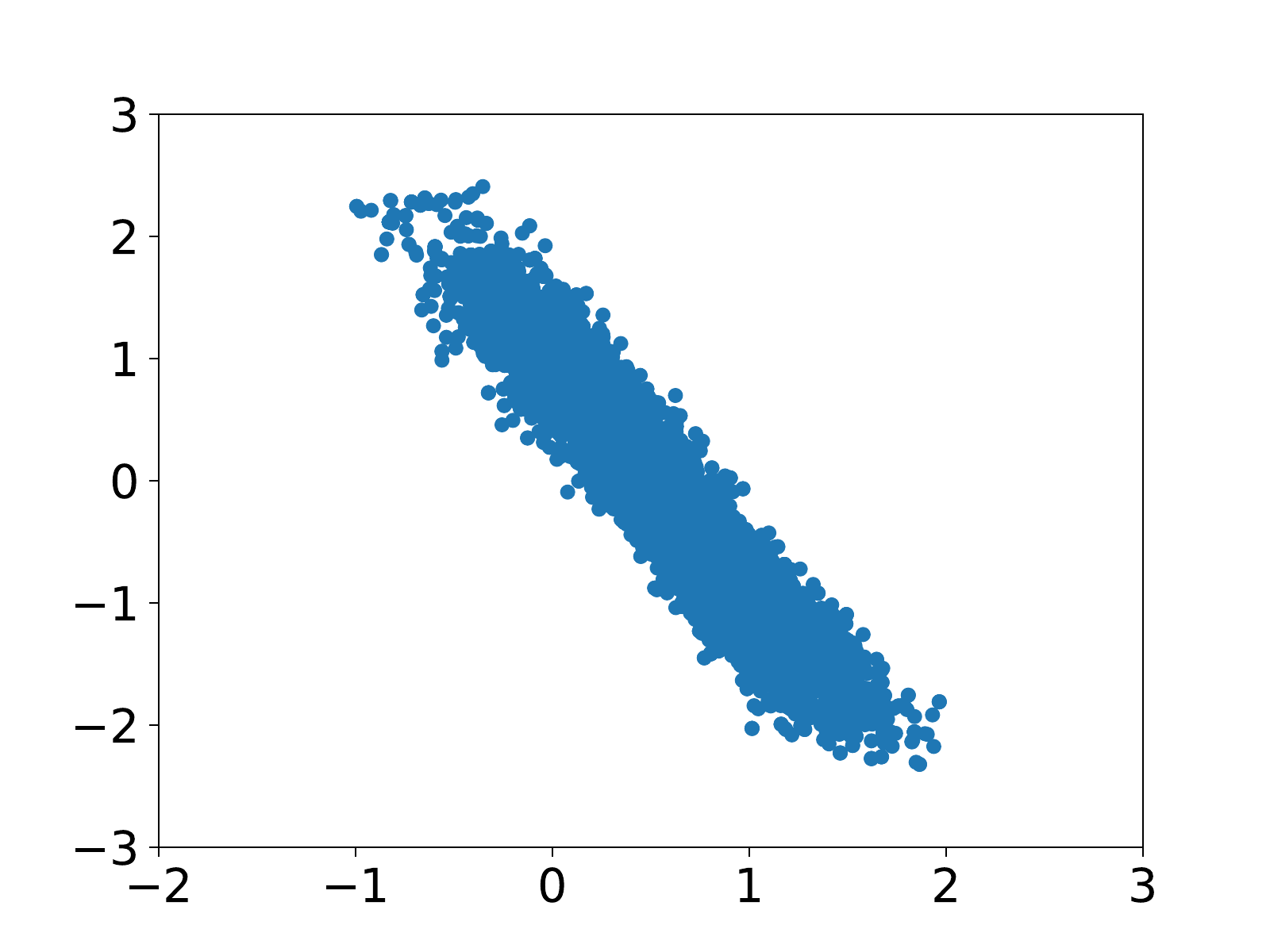}}
            \subfloat[][Penalty]{\includegraphics[width=4.4cm]{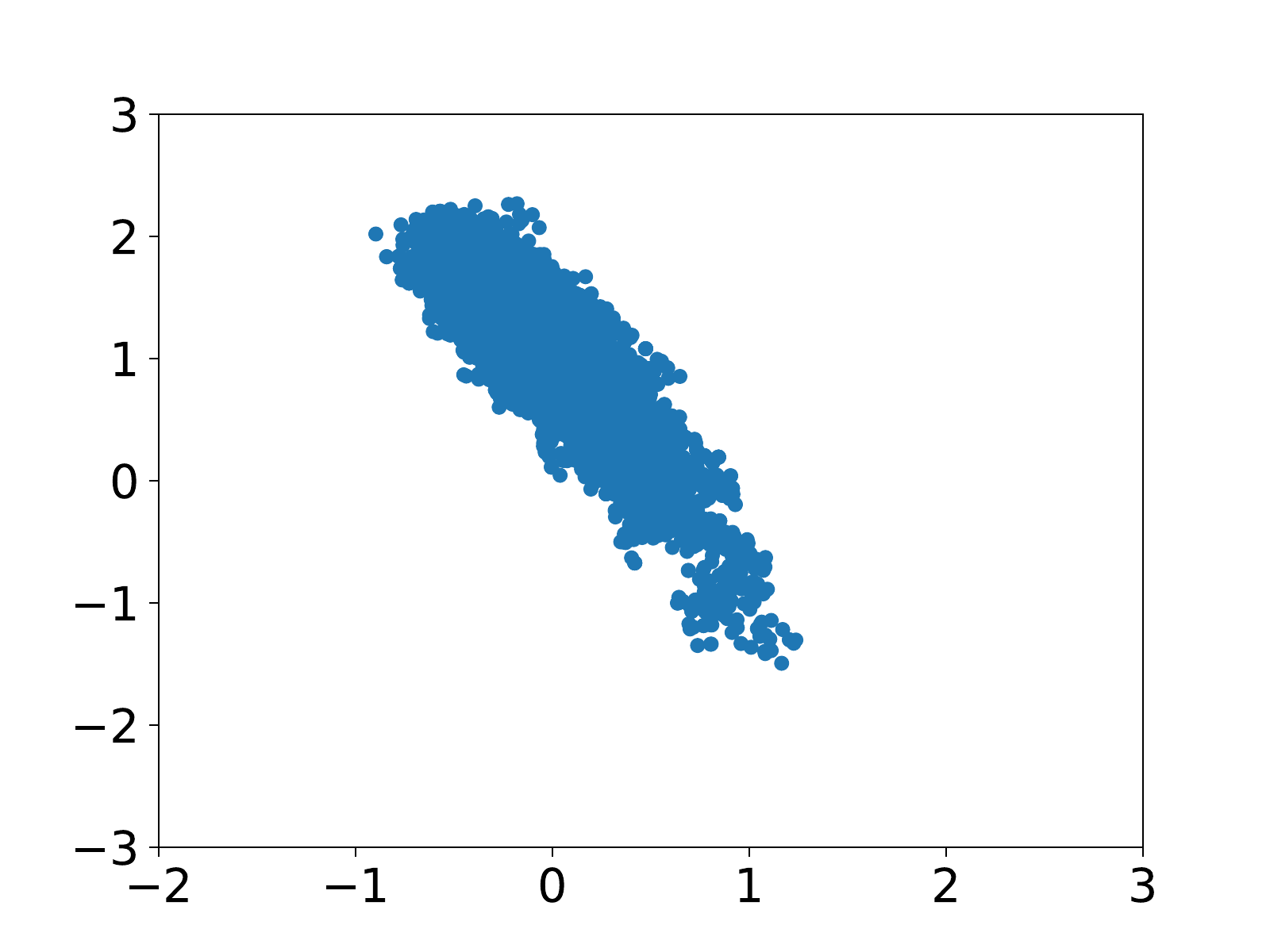}}
            \subfloat[][Barker]{\includegraphics[width=4.4cm]{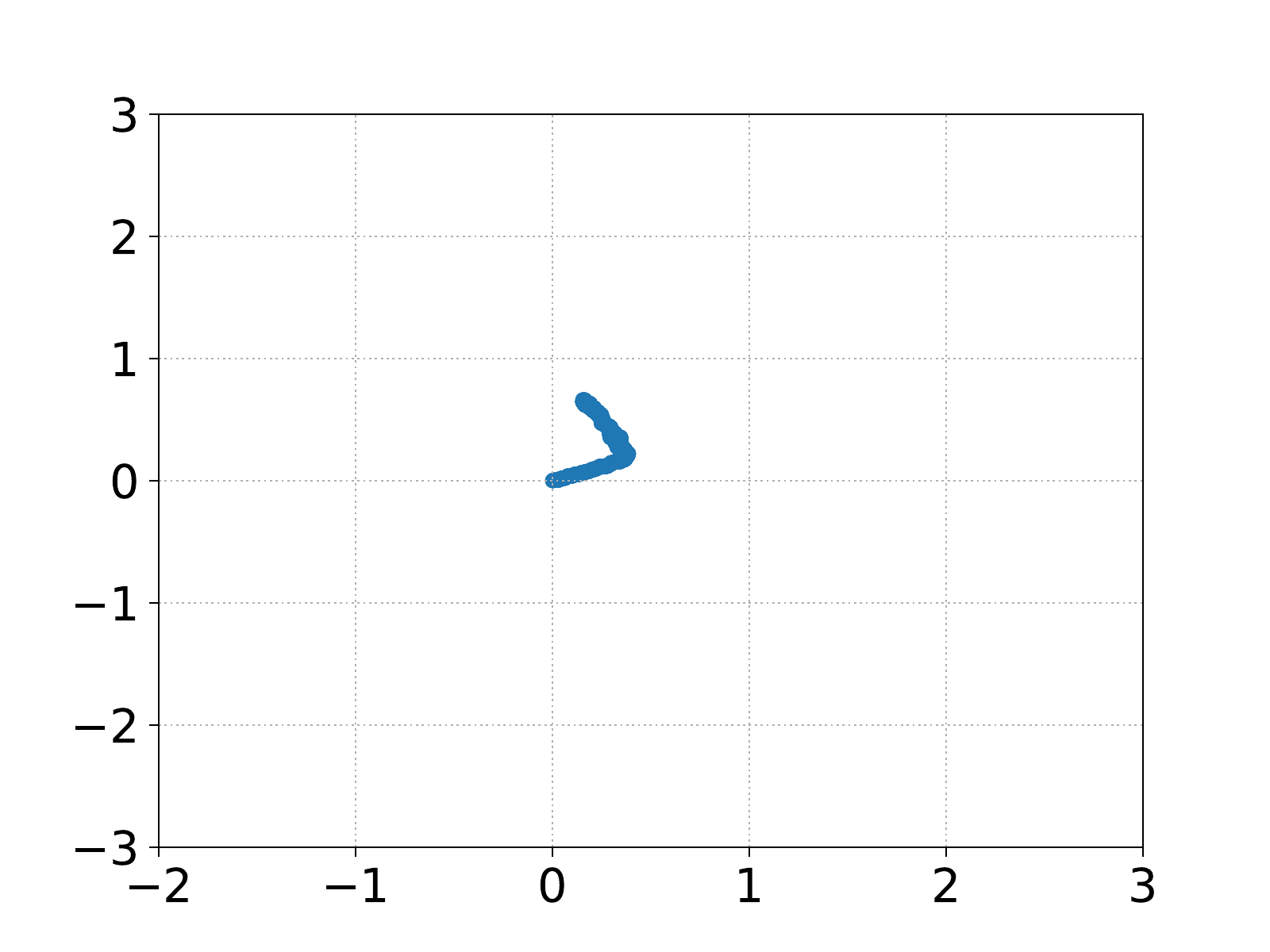}}\\
            \subfloat[][DP-fast MH (full-batch)]{\includegraphics[width=4.4cm]{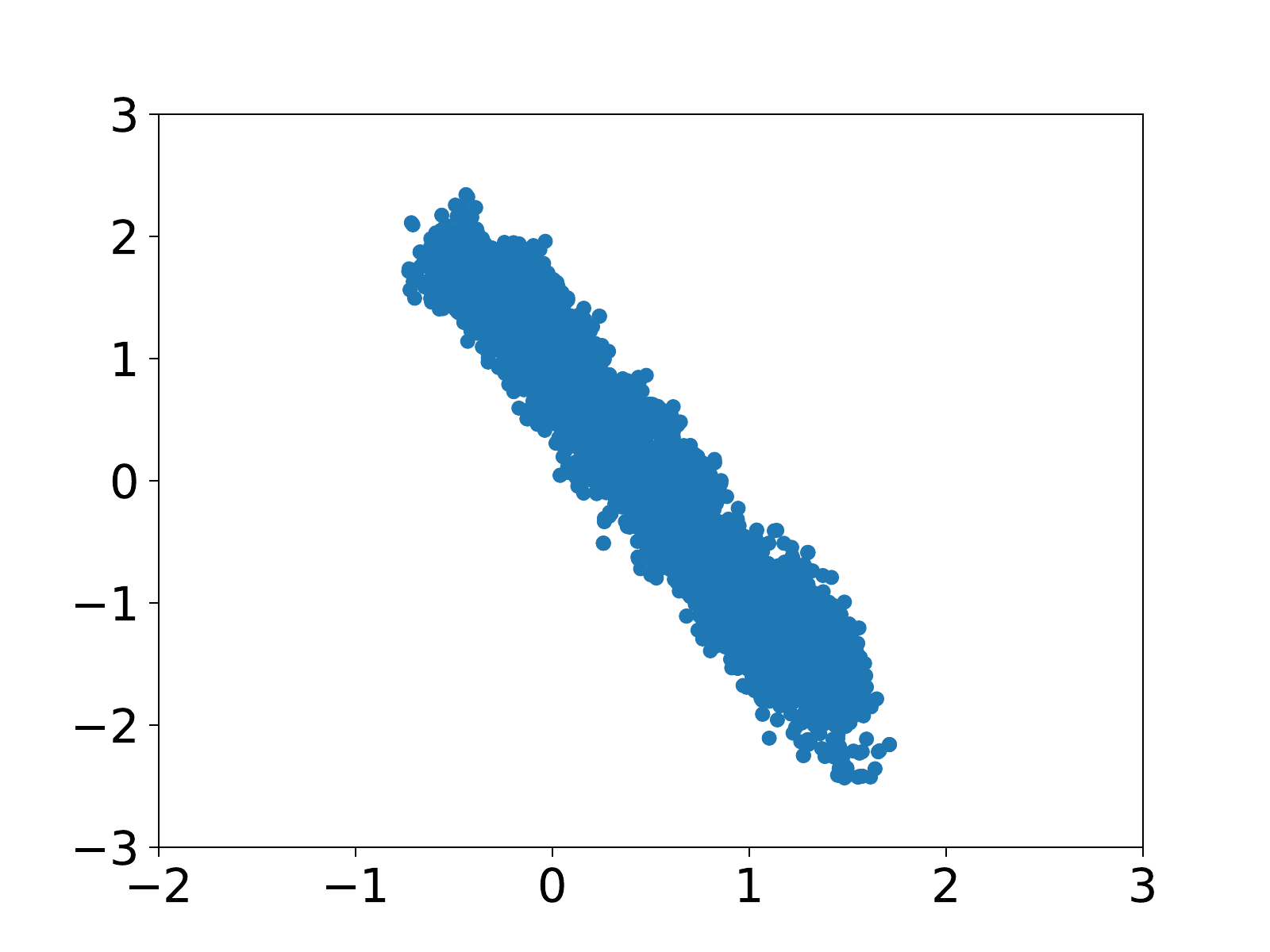}}
            \subfloat[][DP-fast MH]{\includegraphics[width=4.4cm]{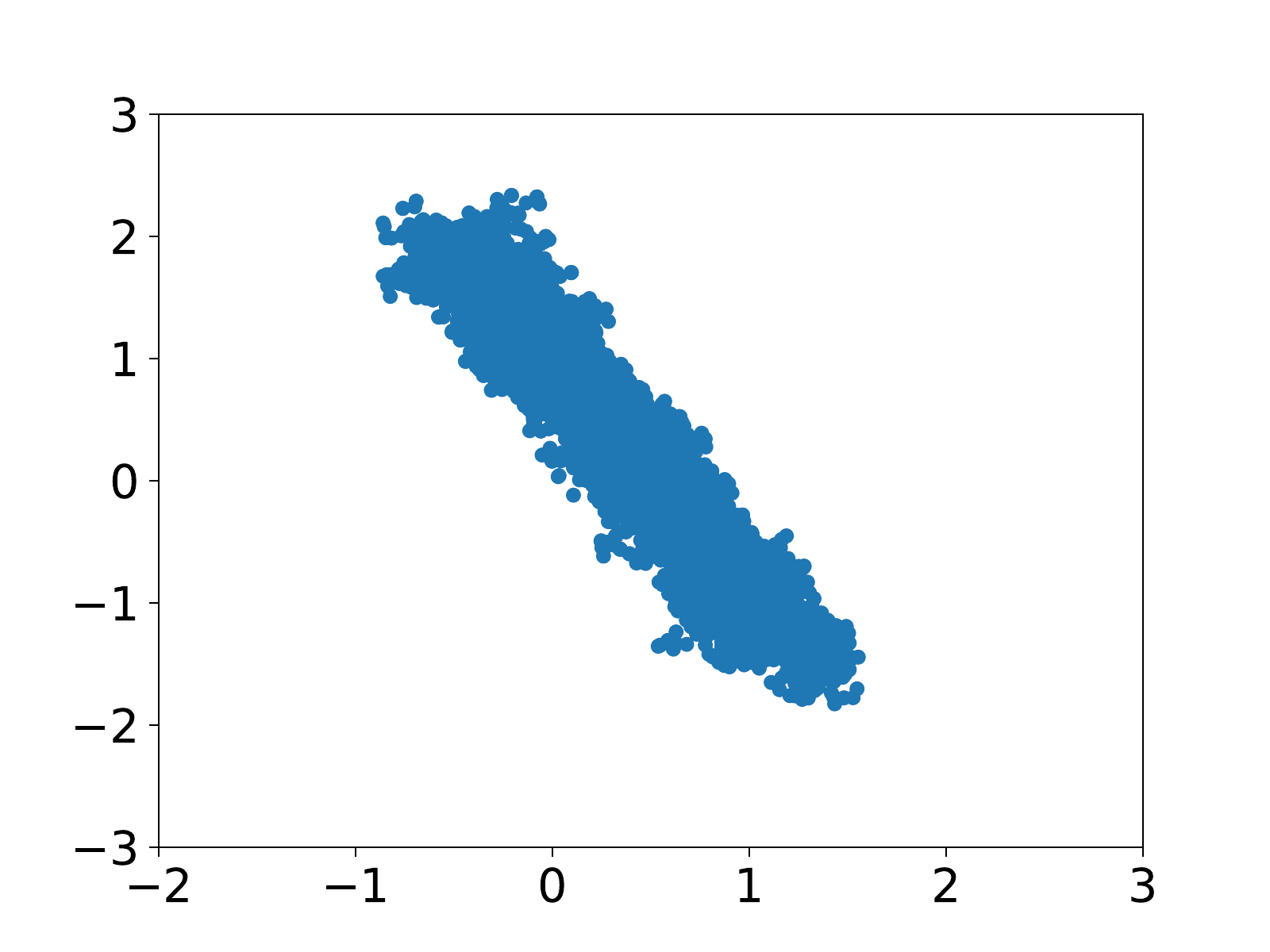}}
          \caption{Density estimation of Gaussian mixtures after $10^{4}$ steps. The estimations of DP-fast MH and its full-bath version are close to the ground truth whereas previous methods are far away from the target distribution. }
	     \label{fig:mog-density}   
		\end{figure*}

\begin{figure*}
	\vspace{-0mm}\centering
	\begin{tabular}{ccccc}		
		\hspace{-4mm}
		\includegraphics[width=4.4cm]{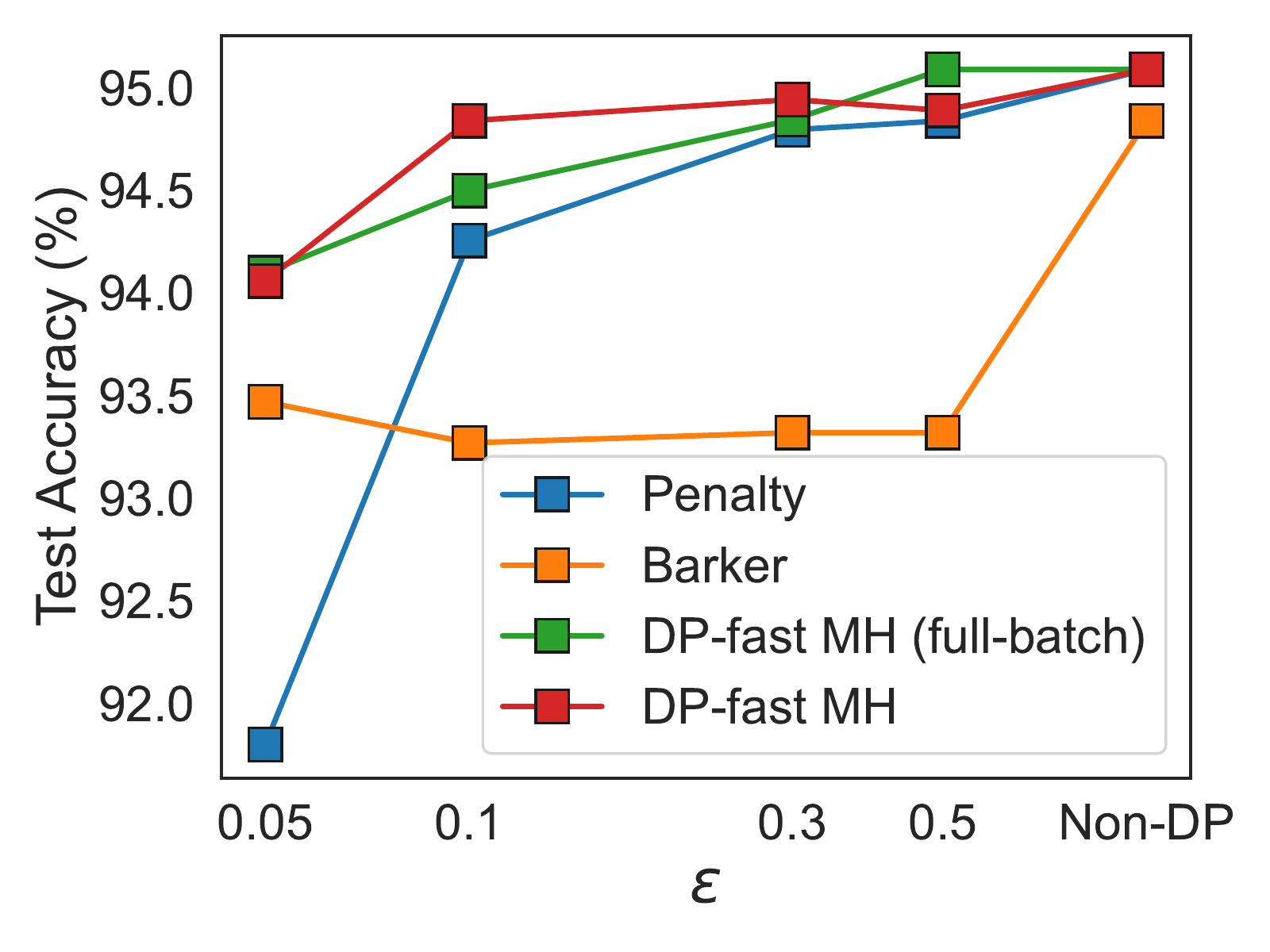}  &
        \hspace{-4mm}
        \includegraphics[width=4.4cm]{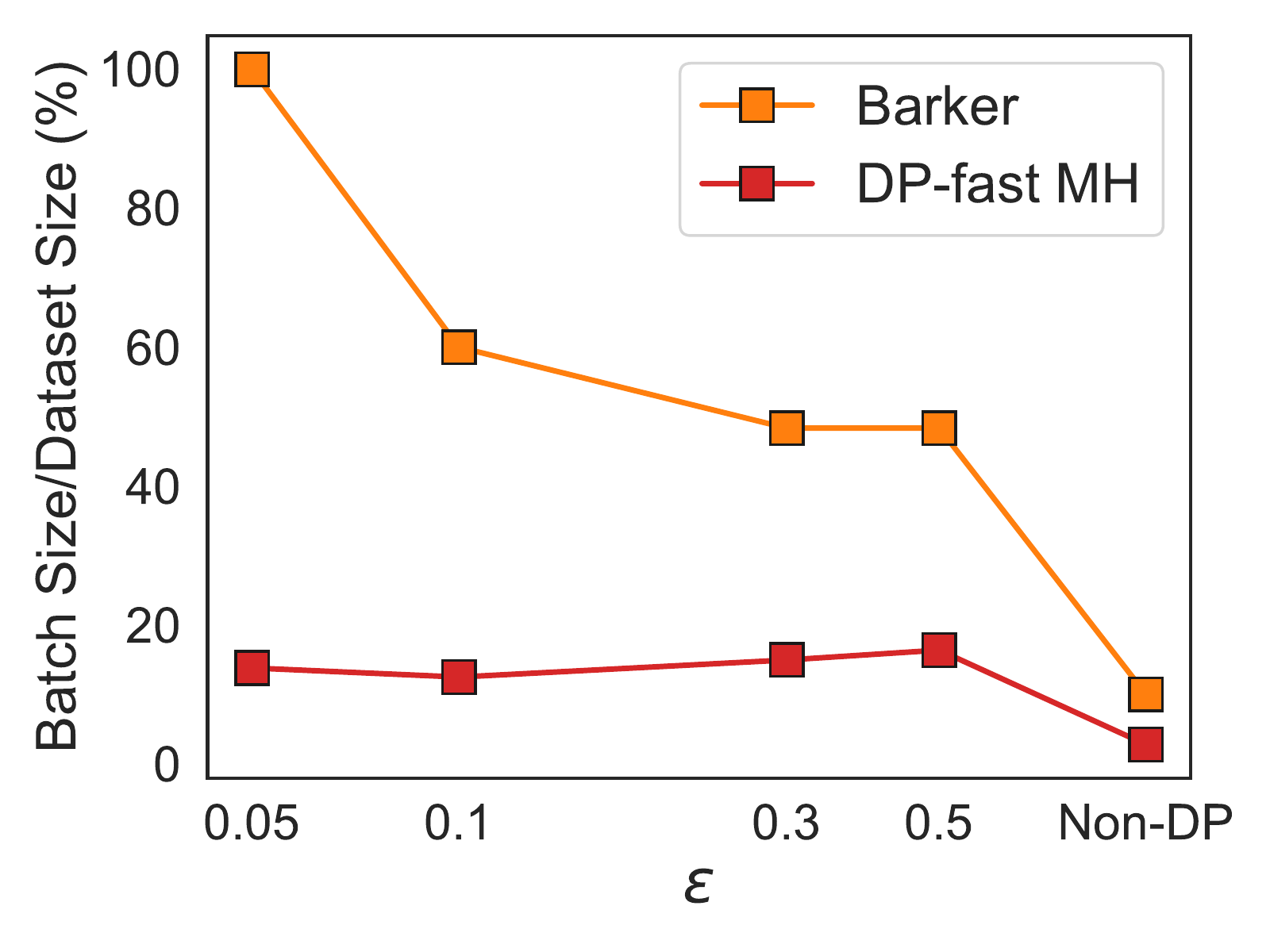}  &
        \hspace{-4mm}
        \includegraphics[width=4.4cm]{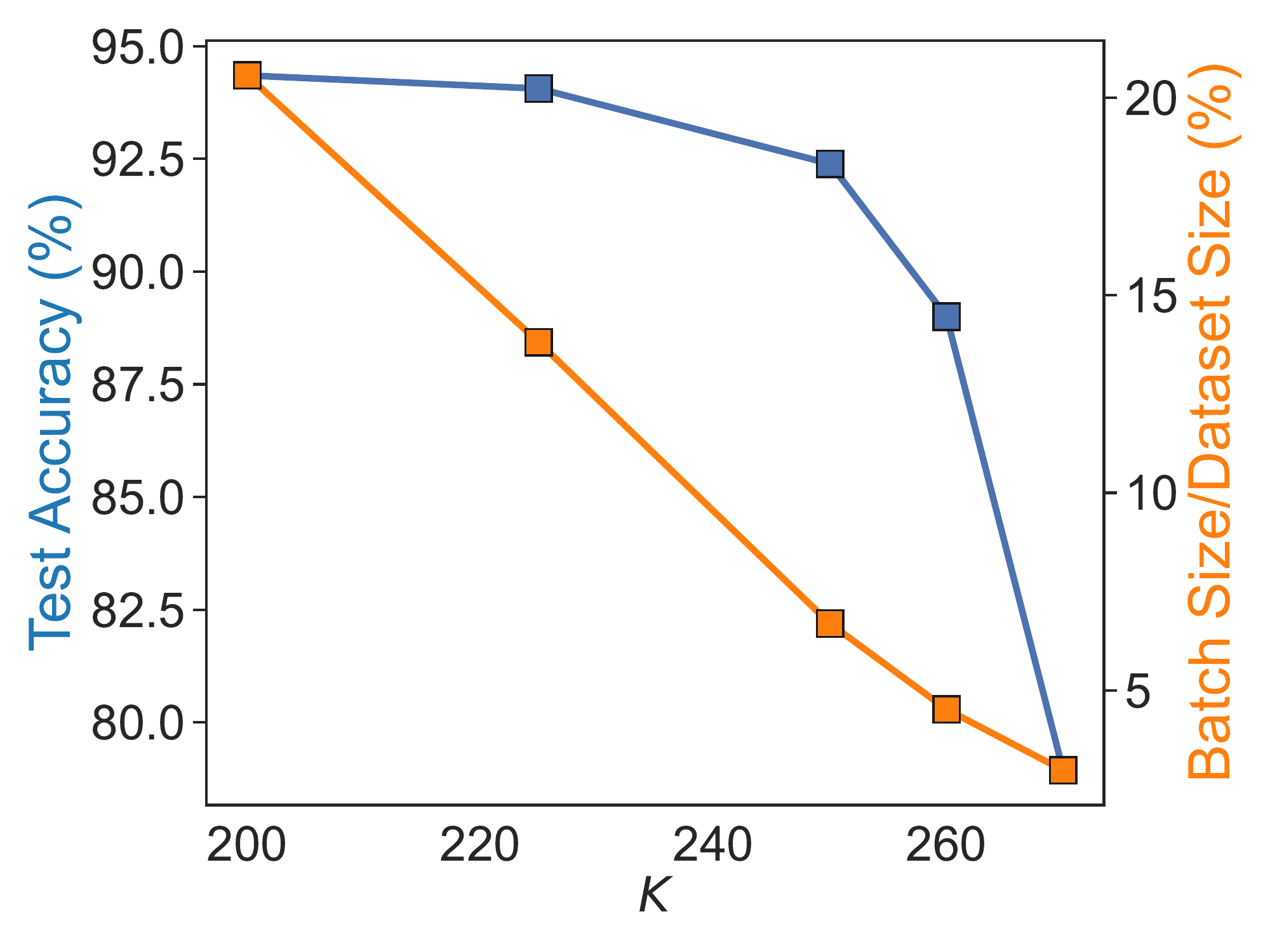}
        \\
		(a)&(b)&(c) 	\\

	\end{tabular}
	\caption{MNIST logistic regression. (a) Test accuracy vs. privacy. (b) Batch size vs. privacy. (c) Effect of hyperparameters $K$. DP-fast MH achieves the highest accuracy with the smallest batch size under all privacy levels.}
	\label{fig:logistic}
\end{figure*}

\subsection{Comparison with Other Private Algorithms}\label{sec:mog}
We compare DP-fast MH with the existing DP MH algorithms: the penalty algorithm~\citep{yildirim2019exact} which adds Gaussian noise to the energy difference and uses full-batch data in every iteration; and the Barker method~\citep{heikkila2019differentially} which uses the Barker acceptance test~\citep{barker1965monte} based on a minibatch approximation of the energy difference. We additionally show the results of the full-batch version of DP-fast MH (i.e. Algorithm~\ref{alg:dp-fastmh} that always instantiates the full-batch MH). Since DP-fast MH trades convergence rate for scalability, DP-fast MH with full-batch data can be regarded as the upper limit of DP-fast MH in terms of performance, if ignoring the computational cost. We also include the results of the corresponding non-DP version for all DP MH methods as references.

\textbf{Truncated Gaussian Mixture.}
We first test DP-fast MH on a two-dimensional truncated Gaussian mixture. Following previous work~\citep{welling2011bayesian, heikkila2019differentially, zhang2020asymptotically}, the data is generated as follows 
\[
x_i \sim \frac{1}{2}\mathcal{N}(\theta_1, \sigma_x^2) + \frac{1}{2}\mathcal{N}(\theta_1+\theta_2, \sigma_x^2)
\]
where $\theta_1 = 0, \theta_2 = 1$ and $\sigma^2 = 2$. The posterior $\theta$ has two modes at $(\theta_1, \theta_2)=(0, 1)$ and $(\theta_1, \theta_2)=(1, -1)$. We generate $N=50000$ samples and use a temperature $500$.

From Figure~\ref{fig:mog}(a), we can see that DP-fast MH and DP-fast MH (full-batch) have better estimation accuracy than previous methods under all privacy levels. Compared to the penalty algorithm, DP-fast MH (full-batch) takes advantage of privacy for free, which greatly improves the convergence of the chain. DP-fast MH (i.e. the minibatch version) has comparable performance with the full batch version, but drastically reduces the cost per step (see Figure(b)). We found that the Barker method is unable to converge to the target distribution because of using a very small step size to satisfy its assumptions\footnote{In the Gaussian mixture experiment in ~\citet{heikkila2019differentially}, the barker method has used a pre-trained solution from variational inference as its initialization. Here, we train all methods from scratch.}. Besides, these results verify our theoretical results on the privacy-efficiency trade-off in Theorem~\ref{thm.convergence-rate}, which shows the efficiency (i.e. the convergence rate) becomes worse when the privacy level increases.

From Figure~\ref{fig:mog}(b), we observe that DP-fast MH achieves a much better posterior approximation with a much smaller batch size compared with the existing minibatch method, Barker method. (Note that we do not include the Penalty method because it is a full-batch method.) This is because DP-fast MH inherits the nice properties from the non-DP baseline TunaMH, which is proved to have an asymptotically optimal batch size. These results also verify our theoretical results on the privacy-scalability trade-off in Theorem~\ref{thm.convergence-rate}, which shows the batch size needs to be larger to keep a high efficiency when the privacy level increases.

Figure~\ref{fig:mog}(c)\&(d) show the convergence with respect to the number of iterations and the wall-clock time under $\epsilon=0.05$ respectively. We see that DP-fast MH converges the fastest, since it converges after 4000 iterations (or 15 seconds),
whereas the penalty algorithm has not fully converged after $10^4$ iterations (or 20 seconds), and the Barker method still has large KL divergence after $10^4$ iterations. Besides, DP-fast MH (full-batch) is also faster than the penalty algorithm, showing the benefits of utilizing the intrinsic randomness in MH to achieve privacy for free.

To illustrate the quality of estimation, we also visualize the density estimate after $10^4$ steps in Figure~\ref{fig:mog-density}. The estimated densities from both DP-fast MH and DP-fast MH (full-batch) are very close to the true posterior distribution (see Figure~\ref{fig:mog-density}(d)\&(e)), while the penalty algorithm cannot estimate the relative weights of the two modes accurately and the Barker method completely failed. These results verify the correct convergence of DP-fast MH and show that under the same privacy budget, DP-fast MH approximates the target distribution significantly better than existing methods.

\textbf{Logistic Regression on MNIST Data.}
Next, we apply DP-fast MH to logistic regression on the MNIST image dataset. Following~\citet{zhang2020asymptotically}, the task is to classify 7s and 9s using the first 50 principal components as features. From Figure~\ref{fig:logistic}(a), we again find that minibatch and full-batch DP-fast MH both outperform previous methods significantly on test accuracy under all privacy levels, demonstrating that DP-fast MH achieves a better privacy-utility trade-off due to leveraging the inherent randomness in the MH algorithm. Similar to Figure~\ref{fig:mog}(b), Figure~\ref{fig:logistic}(b) also demonstrates the better scalability of DP-fast MH compared to the previous minibatch method. When the privacy parameter $\eps$ goes smaller, the previous method has to use almost the entire dataset while our method only requires less than $20\%$ data to obtain even better accuracy. 

\textbf{DP-fast MH vs. DP-fast MH (Full-Batch).}
DP-fast MH can be used with minibatch or full-batch of data. We now discuss their usage scenarios. The minibatch version of DP-fast MH will be preferred when 
the dataset size is large or the computational resource is limited. The full-batch version of DP-fast MH will be preferred if the dataset size is relatively small such that the number of convergence steps, rather than the cost per step, is the main concern. 

\vspace{-1mm}
\subsection{Choice of $K$}\label{sec:choicek}
DP-fast MH only introduces one new hyperparameter $K$, which is the minibatch upper bound, compared to the baseline non-DP method TunaMH. We now discuss how to choose this hyperparameter $K$ in practice. Given a target privacy budget, $K$ has two effects: (1) it affects the computation efficiency, since if the random batch size is greater than $K$, the standard MH will be invoked (as shown in Algorithm~\ref{alg:dp-fastmh}), which requires computation on the entire data. So the computational cost increases as $K$ decreases. (2)
Meanwhile, $K$ affects the convergence rate of the chain. As shown in Theorem~\ref{thm.convergence-rate}, a smaller $K$ results in a larger spectral gap upper bound, thus a faster convergence rate. 

We use the logistic regression task to investigate the performance under different choices of $K$ with privacy parameter (in the Gaussian mechanism) $\eps=0.05$. Figure~\ref{fig:logistic}(c) verifies our theoretical analysis on $K$, since the test accuracy and the batch size both increase when $K$ decreases. We recommend setting $K$ to be around $\eps C/(6\max_ic_i)$ to achieve a good trade-off between efficiency and scalability, since the privacy parameter in the Gaussian mechanism of the minibatch MH (i.e. $\eps C/(6K\max_ic_i$)) is roughly a constant under this value of $K$.

\vspace{-2mm}
\section{Conclusions and Future Work}
We present the first exact minibatch DP MCMC algorithm, DP-fast MH, for private large-scale Bayesian inference. Our method is developed by using Gaussian mechanisms in a way that maintains the reversibility of the Markov chain. Furthermore, we leverage the intrinsic randomness in the Metropolis-Hastings algorithm to achieve privacy for free under uncertain cases. We provide privacy, asymptotic convergence, and convergence rate guarantees, revealing a new three-way trade-off among privacy, scalability, and efficiency in Bayesian inference. We demonstrate on Gaussian mixture and logistic regression tasks that DP-fast MH significantly outperforms previous methods, indicating that our method has a much better privacy-utility trade-off. 
Moreover, DP-fast MH is simple to implement and can be broadly applied as
a drop-in replacement for the standard MH when the target distribution has a bounded energy function.

While Bayesian inference is one of the major tools for analyzing data, a general inference algorithm that can guarantee privacy while retaining accuracy, scalability, and efficiency was largely unexplored. We believe this work fills an important gap and will accelerate the practical use of Bayesian inference for privacy-critical applications.

\section*{Acknowledgement}
W.Z. is supported by a Computing Innovation Fellowship from the Computing Research Association (CRA) and the Computing Community Consortium (CCC), and the BU Census grant.

\bibliography{references}
\bibliographystyle{icml2023}
\renewcommand\appendix{\par
    \setcounter{section}{0}
    \setcounter{subsection}{0}
    \gdef\thesection{ \Alph{section}}}
\newpage 
\appendix
\onecolumn

\section{Proof of Lemma \ref{lemma.sen}}

\begin{proof}
Fix any two neighboring datasets $D, D'$ that differ on index $i$. We denote the energy difference for the minibatch as $\ell_1$ and $\ell'_1$, and the energy difference for the entire data as $\ell_2$ and $\ell'_2$. By the definition of $\ell_1$, we have $\ell'_1=\ell_1+ \Delta$, where 
\begin{align*}
\Delta=&2\left(\operatorname{artanh}\left(
        \frac{C \left(U_i(\theta|x_i') - U_i(\theta'|x_i') \right)}{c_i' (2 \lambda + C' M(\theta,\theta'))} 
    \right) \right). \\
        &- \left. \operatorname{artanh}\left(
        \frac{C \left(U_i(\theta|x_i) - U_i(\theta'|x_i) \right)}{c_i (2 \lambda + C M(\theta,\theta'))} 
    \right) \right).
\end{align*}

Because $\operatorname{artanh} x =\frac{1}{2}\log \frac{1+x}{1-x}$, the maximum $\max_{x_i} \operatorname{artanh}\left(
        \frac{C \left(U_i(\theta) - U_i(\theta') \right)}{c_i (2 \lambda + C M(\theta,\theta'))} 
    \right)$ must be obtained when $U_i(\theta|x_i)>U_i(\theta'|x_i)$, otherwise the $\log$ term is negative. Since $\frac{1+x}{1-x}$ is monotonically increasing when $0\le x<1$, and by Assumption \ref{assump} that $|U_i(\theta) - U_i(\theta')|\le c_i M(\theta,\theta')$,
        the maximum is reached when $U_i(\theta) - U_i(\theta') = c_i M(\theta,\theta')$.  Thus, we have
\begin{align*}
\MoveEqLeft \max_{x_i} \operatorname{artanh}\left(
        \frac{C \left(U_i(\theta) - U_i(\theta') \right)}{c_i (2 \lambda + C M(\theta,\theta'))} 
    \right) \notag \\
        &\le \frac{1}{2} \log \left(\frac{2\lambda c_i + 2c_i CM}{2\lambda c_i}\right) \\
        &=\frac{1}{2} \log (1+\frac{CM}{\lambda }), \notag
\end{align*}

Similarly, $\min_{x_i} \operatorname{artanh}\left(
        \frac{C \left(U_i(\theta) - U_i(\theta') \right)}{c_i (2 \lambda + C M(\theta,\theta'))} 
    \right)$ is bounded below by $- \frac{1}{2} \log (1+\frac{CM}{\lambda })$. Hence,
$\Delta\le \Delta (\ell_1):=2 \log (1+\frac{CM}{\lambda }).$

For $\ell_2$, we have
$$\ell'_2=\ell_2+\Delta,$$
where $\Delta=U_i(\theta|x'_i) - U_i(\theta'|x'_i)-U_i(\theta|x_i) + U_i(\theta'|x_i)$. By by Assumption \ref{assump}, we have that $\Delta\le \Delta(\ell_2) := 2\max_i c_i M(\theta,\theta')$.

\end{proof}

\section{Proof of Theorem \ref{thm.privacy}}\label{app.privacyproof}
\begin{proof}

\textbf{Gaussian mechanism.} For the Gaussian noise that we use, if we choose $\sigma$ in Algorithm \ref{alg:dp-fastmh} to be $\sqrt{2\log\frac{1.25}{\delta}}/\eps$, then each step is $(\eps, \delta)$-differentially private with respect to the minibatch. The Gaussian noise is also scaled with the sensitivity $\delta(\ell)$ which depends on the current $\theta$ and the proposed $\theta'$.

\textbf{Sampling (with replacement) with non-uniform probabilities.} 
When forming a minibatch, we sample index $i$ with probability $p_i=c_i/C$.
Let $D$ and $D'$ be neighboring datasets of size $n$, and we assume that without loss of generality they differ in the last row. The worst case would be the last row has the highest probability of being sampled into the minibatch. Let $p=\max_i c_i/C$. We note that after the Poisson sampling step, for a given sample $i_b$,  the additional step of determining whether to add $i_b$ to the batch has a privacy amplification effect. Specifically, if the subsequent mechanism is $(\epsilon, \delta)$-differentially private, this step will result in $(\epsilon', \delta')$-differential privacy, where $\epsilon' = \log(1 + \beta(\exp(\epsilon) - 1))$ and $\delta' = \beta \delta$, with $\beta = \max_{i_b} \Pr[\text{add }i_b]$. This follows directly from Theorem 1 in \citep{li2012sampling}. Note that $\epsilon' \leq \epsilon$ and $\delta' \leq \delta$. We take $(\epsilon, \delta)$-differential privacy as a conservative privacy guarantee, since $\beta$ depends on the data and is upper-bounded by 1.

For a given batch size $B$, let $\mathcal{M}_1$ denote the Poisson sampling step and 
let $\mathcal{M}_2$ be the $(\eps, \delta)$-DP  mechanism on a minibatch. 
Let $T$ be a random variable denoting the multiset of indices sampled by $\mathcal{M}_1$, and let $\ell(T)$ be the multiplicity of index $n$ in $T$.
Fixing a subset of possible log scale of the acceptance probability. For each $k=0,1,\ldots, B$, let
\begin{align*}
p_k&=\Pr[\ell(T)=k |B]= {B \choose k} p^k(1-p)^{B-k}\\
q_k&=\Pr[\mathcal{M}_2(D|_T)\in \mathcal{S}|\ell(T)=k, B] \\
q'_k&=\Pr[\mathcal{M}_2(D'|_T)\in \mathcal{S}|\ell(T)=k, B]
\end{align*}

By privacy of $\mathcal{M}_2$, we have $q_k\le \exp(\eps)q_{k-1}+\delta$, so $q_k\le \exp(k\eps)q_0+\frac{\exp(k\eps)-1}{\exp(\eps)-1}\delta$. Hence,
\begin{align}
\Pr[\mathcal{M}_1\circ \mathcal{M}_2 (D)\in \mathcal{S}|B] &= \sum_{k=0}^B p_kq_k \notag \\
&\le \sum_{k=0}^B {B \choose k} p^k(1-p)^{B-k} \left( \exp(k\eps)q_0+\frac{\exp(k\eps)-1}{\exp(\eps)-1}\delta \right) \notag \\
&=q_0(1-p+pe^\eps)^B+\frac{\delta}{e^\eps-1}[(1-p+pe^\eps)^B-1]. \label{eq.priv1}
\end{align}
Similarly, we have 
\begin{equation}
\Pr[\mathcal{M}_1\circ \mathcal{M}_2 (D')\in \mathcal{S}|B]\ge q_0(1-p+p\exp(-\eps))^B-\frac{\delta}{\exp(-\eps)-1}[(1-p+p\exp(-\eps))^B-1].
\end{equation}

Then we account for the randomness of the batch size $B$. To simplify the notation, we denote $\chi C^2M^2(\theta,\theta')  + CM(\theta,\theta')$ as $\lambda$, we have
\begin{align}
\Pr[\mathcal{M}_1\circ \mathcal{M}_2 (D)\in \mathcal{S}] &\le \sum_{B=0}^K \left( q_0(1-p+pe^\eps)^B+\frac{\delta}{e^\eps-1}[(1-p+pe^\eps)^B-1] \right) \frac{\lambda^B \exp(-\lambda)}{B!} \notag \\
&\le  q_0(1-p+pe^\eps)^K+\frac{\delta}{e^\eps-1}[(1-p+pe^\eps)^K-1] \label{eq.priv2}
\end{align}

Similarly, we also have that 
\begin{align}\label{eq.priv3}
\Pr[\mathcal{M}_1\circ \mathcal{M}_2 (D')\in \mathcal{S}] &\ge \sum_{k=0}^K \left( q_0(1-p+pe^{-\eps})^B-\frac{\delta}{e^{-\eps}-1}[(1-p+pe^{-\eps})^B-1] \right) \frac{\lambda^B \exp(-\lambda)}{B!}\\
& \ge q_0(1-p+pe^{-\eps})^K
\end{align}

Combining inequalities \eqref{eq.priv2} and \eqref{eq.priv3}, we obtain 
\begin{align*}
    \Pr [\mathcal{M}_1\circ \mathcal{M}_2 (D)\in \mathcal{S}]&\le (\frac{1-p+p\exp(\eps)}{1-p+p\exp(-\eps)})^K 
\\&\Pr[\mathcal{M}_1\circ \mathcal{M}_2 (D')\in \mathcal{S}]
\\&+ \frac{\delta}{\exp(\eps)-1}[(1-p+p\exp(\eps))^K-1]  .
\end{align*}

Then $\mathcal{M}_1\circ \mathcal{M}_2$ is $(\eps', \delta')$-differentially private for
\begin{equation*}
\eps'\le K\log \frac{1-p+p\exp(\eps)}{1-p+p \exp(-\eps)}) \le  6Kp\eps,
\end{equation*}

and 
\begin{equation*}
\delta'\le 2Kp\delta.
\end{equation*}

\textbf{Account the possibility of using standard MH.}
With probability $q=\sum_{k=K+1}^\infty \frac{\lambda^k \exp(-\lambda)}{k!}$, our algorithm will invoke the private standard MH algorithm. We denote our algorithm per iteration be $\tilde{\mathcal{M}}$, and the private standard MH be $\mathcal{M}_3$. Then we have
\begin{align*}
\Pr[\tilde{\mathcal{M}}(D)\in \mathcal{S}] & = q \Pr [\mathcal{M}_1\circ \mathcal{M}_2 (D)\in \mathcal{S}]+(1-q)\Pr[\mathcal{M}_3(D)\in \mathcal{S}] \\
&\le q (\exp(\eps') \Pr[\mathcal{M}_1\circ \mathcal{M}_2 (D')\in \mathcal{S}]+\delta') + (1-q)(\exp(\eps') \Pr[\mathcal{M}_3(D') \in \mathcal{S}]+\delta')\\
&=\exp(\eps')\Pr[\tilde{\mathcal{M}}(D')\in \mathcal{S}]+\delta'.
\end{align*}

\textbf{Composition over $T$ iterations.}
Last, we obtain the total privacy loss using the advanced composition over $T$ iterations. Algorithm \ref{alg:dp-fastmh} is $(\eps, T\delta'+\delta)$-differentially private for $\eps=\sqrt{2T\log(1/\delta)}\eps'+T\eps'(e^{\eps'}-1)$. We remark that we can obtain a tighter composition bound by using GDP \cite{dong2019gaussian}.

\end{proof}

\section{Proof of Theorem~\ref{thm.asymptotic}}\label{app:asym-convergence}

\begin{proof}
We prove that DP-fast MH is $\pi$-reversible. Let the transition operator be
\[
T_{\text{DP-fast MH}}(\theta,\theta') = q(\theta'|\theta)a_{\text{DP-fast MH}}(\theta,\theta')
\]
where $a_{\text{DP-fast MH}}(\theta,\theta')$ denotes the acceptance probability of DP-fast MH.
DP-fast MH decides whether to use TunaMH or standard MH based on the minibatch size $B$ and the size upper bound $K$, and it also decides whether to add Gaussian noise or not based on $\Delta(l_1),\Delta(l_2)$ and $\epsilon$. Specifically, we have
\begin{align*}
    a_{\text{DP-fast MH}}(\theta,\theta') &= \mathbb{I}(B<K)\left(\mathbb{I}(\Delta(l_1)>\frac{\eps C}{6 K\max_i c_i})a_{\text{DP-TunaMH}}+\mathbb{I}(\Delta(l_1)\le\frac{\eps C}{6 K\max_i c_i}) a_{\text{TunaMH}}(\theta,\theta')\right)\\
&\hspace{-5em}+\mathbb{I}(B\ge K)\left(\mathbb{I}(\Delta(l_2)>\eps)a_{\text{DP-MH}}(\theta,\theta')+\mathbb{I}(\Delta(l_2)\le \eps )a_{\text{MH}}(\theta,\theta')\right).
\end{align*}
where $a_{\text{DP-TunaMH}}, a_{\text{DP-MH}}$ denote the acceptance probability of TunaMH and MH with added Gaussian noise.

Our key observation is that all indicator functions are symmetric in $\theta$ and $\theta'$ when $M(\theta,\theta')$ is symmetric in $\theta$ and $\theta'$ . Besides, following the analysis of the penalty algorithm~\citep{ceperley1999penalty,yildirim2019exact}, DP-TunaMH and DP-MH are both reversible. Combining all these results, we can show that 
\begin{align*}
    \pi(\theta)T_{\text{DP-fast MH}}(\theta,\theta')&=\pi(\theta)q(\theta'|\theta)a_{\text{DP-fast MH}}(\theta,\theta')
\end{align*}
is symmetric in $\theta$ and $\theta'$.
\end{proof}

\section{Proof of Theorem~\ref{thm.convergence-rate}}\label{app:convergence-rate}
\begin{proof}
We consider the acceptance probability ratio
\begin{align*}
   \frac{a_{\text{DP-fast MH}}(\theta,\theta')}{a_{\text{MH}}(\theta,\theta')} 
   &\ge
   \min\left(\frac{a_{\text{DP-TunaMH}}(\theta,\theta')}{a_{\text{MH}}(\theta,\theta')}, \frac{a_{\text{DP-MH}}(\theta,\theta')}{a_{\text{MH}}(\theta,\theta')}\right)\\
   &=
   \min\left(\frac{a_{\text{DP-TunaMH}}(\theta,\theta')}{a_{\text{TunaMH}}(\theta,\theta')}\cdot \frac{a_{\text{TunaMH}}(\theta,\theta')}{a_{\text{MH}}(\theta,\theta')},\frac{a_{\text{DP-MH}}(\theta,\theta')}{a_{\text{MH}}(\theta,\theta')}
   \right)
   \\&\ge
   \min\left(\left(1-\Phi\left(\frac{\sigma^2_1\Delta(l_1)^2}{2}\right)\right) \cdot \exp\left(-\frac{C^2A^2}{\lambda } -2\sqrt{\frac{C^2A^2}{\lambda } \log 2}\right)
   ,
   \left(1-\Phi\left(\frac{\sigma^2_1\Delta(l_1)^2}{2}\right)\right)
   \right)
   ,
\end{align*}
where the last equation is obtained using the results in Theorem 2 in~\citet{zhang2020asymptotically} and Proposition 2 in~\citet{yildirim2019exact}.
By the assumption that $M(\theta,\theta')\le A$, it follows that
\[
\Delta(\ell_1)=2\log \left(1+ \frac{CM(\theta,\theta')}{\lambda}\right)\le 2\log \left(1+ \frac{CA}{\lambda}\right),~~~\Delta(\ell_2)=2\max_i c_i M(\theta,\theta')\le 2A\max_i c_i 
\]

Since $\sigma_1=6K\max_i c_i\sqrt{2\log\frac{2.5 K\max_i c_i}{\delta C}}/(\eps C)$ and $\sigma_2=\sqrt{2\log\frac{1.25}{\delta}}/\eps$, then
\[
\sigma_1^2\Delta(l_1)^2\le \frac{216K^2\max_i c_i^2\log\frac{2.5 K\max_i c_i}{\delta C}\log^2 \left(1+ \frac{CA}{\lambda}\right)}{\eps^2 C^2},~~~\sigma_2^2\Delta(l_2)^2\le \frac{8A^2\max_i c_i^2\log\frac{1.25}{\delta}}{\eps^2}.
\]
The acceptance probability ratio thus has the following bound
\begin{align*}
   &\frac{a_{\text{DP-fast MH}}(\theta,\theta')}{a_{\text{MH}}(\theta,\theta')} \\
   \ge &
   \min\left(\left(1-\Phi\left(\frac{32(\log 2)^2\log\frac{1.25}{\delta}}{\eps^2}\right)\right) \cdot \exp\left(-\frac{C^2A^2}{\lambda } -2\sqrt{\frac{C^2A^2}{\lambda } \log 2}\right)
   ,
   \left(1-\Phi\left(\frac{2C^2A^2\log\frac{1.25}{\delta}}{9K^2\eps^2}\right)\right)
   \right)
   .
\end{align*}
Let $\bar{\gamma}$ denote the
spectral gap of DP-fast MH, and let $\gamma$ denote the spectral gap of the standard MH with the same target
distribution $\pi$ and the proposal distribution $q$. By a Dirichlet form argument, we have
\begin{align*}
   \frac{\bar{\gamma}}{\gamma} 
   &\ge
   \left(1-\Phi\left(\frac{216K^2\max_i c_i^2\log\frac{2.5 K\max_i c_i}{\delta C}\log^2 \left(1+ \frac{CA}{\lambda}\right)}{\eps^2 C^2}\right)\right) \cdot \exp\left(-\frac{C^2A^2}{\lambda } -2\sqrt{\frac{C^2A^2}{\lambda } \log 2}\right)
   .
\end{align*}
\end{proof}

\section{Experimental Details}
\subsection{Truncated Gaussian Mixture}
The data is generated from a mixture of two Gaussians. In order to get the bounds required by all methods, the Gaussian is truncated by setting $x_i\in [-3,3]$. We assume a flat prior $p(\theta)=1$. Then the energy is given by
\[
U_i(\theta) = -\log p(x_i|\theta) 
= \log(2\sqrt{2\pi}\sigma_x) - \log\bigg[\exp\bigg(-\frac{(x_i-\theta_1)^2}{2\sigma_x^2}\bigg) + \exp\bigg(-\frac{(x_i-\theta_1-\theta_2)^2}{2\sigma_x^2}\bigg)\bigg]. 
\]
We can set $M(\theta,\theta') = \norm{\theta-\theta'}_2$ and 
\[
c_i = \sqrt{\bigg(\frac{2\Abs{x_i}+9}{\sigma_x^2}\bigg)^2 + \bigg(\frac{\Abs{x_i}+6}{\sigma_x^2}\bigg)^2}.
\]

We tune the stepsize of each method to reach the acceptance rate $60\%$ and set $K$ to be around $\eps C/\max_ic_i$. The hyperparameters in baseline methods are set following their original papers~\citep{yildirim2019exact,heikkila2019differentially}.

\subsubsection{Additional Experimental Results}
\begin{wrapfigure}{R}{.35\columnwidth}
    \centering
    \includegraphics[width=.35\columnwidth]{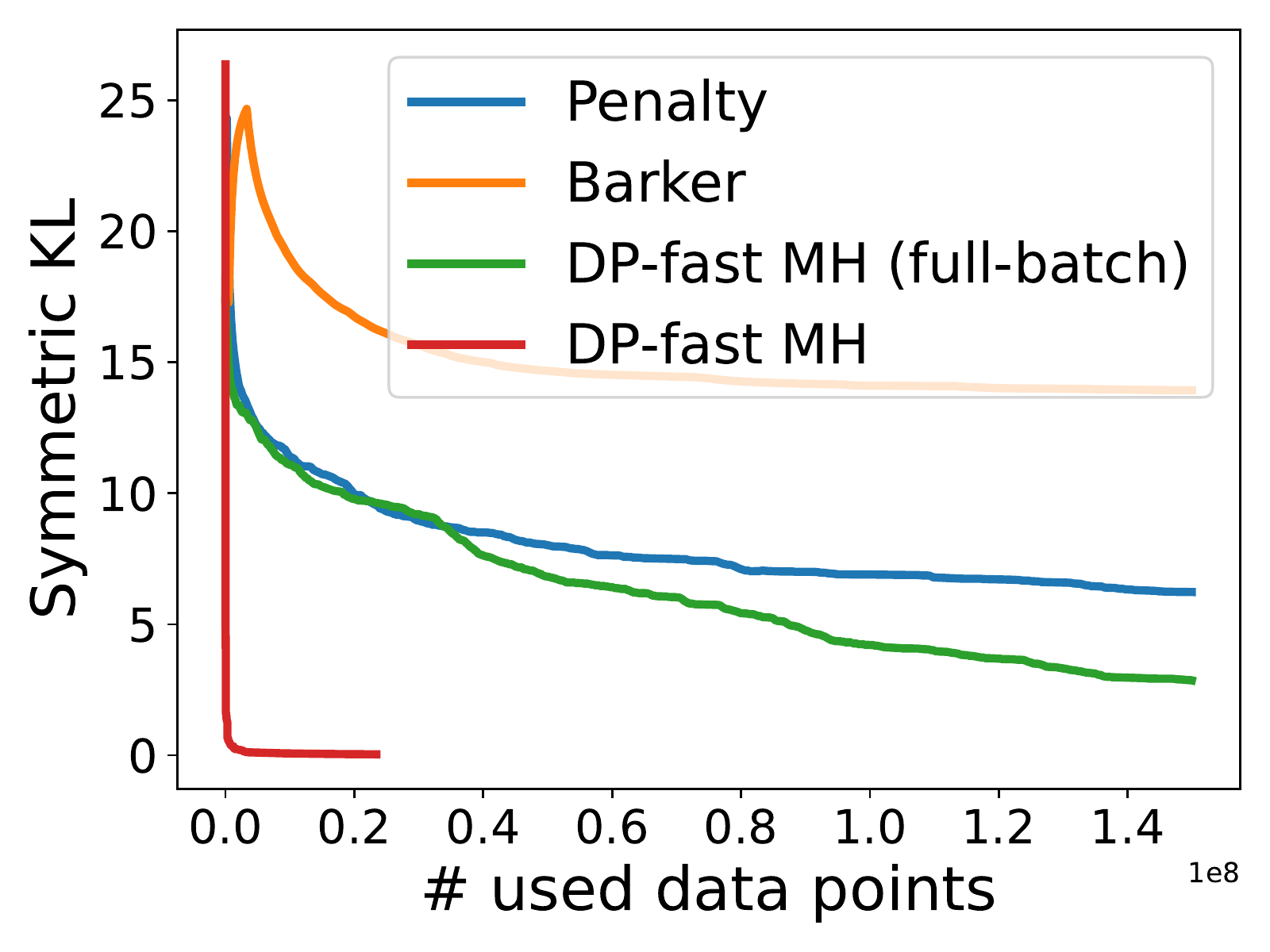}
    \caption{Symmetric KL vs. the number of used data points when privacy $\epsilon=0.05$.}
    \label{fig:mog-time}
\end{wrapfigure}

We additionally show symmetric KL vs. the number of used data points in Figure~\ref{fig:mog-time}, which does not depend on the implementation details.  From this figure, we can clearly see that both minibatch and full-batch DP-fast MH converge significantly faster than previous methods.

\subsection{Logistic Regression on MNIST}

MNIST with only 7s and 9s images contains 12214 training data and 2037 test data. Let $h$ be the sigmoid function. Let the label $y_i\in \{0, 1\}$, then the model in logistic regression (LR) is 
\[
p(y_i = 1) = h(\theta^\intercal x_i) = \frac{1}{1 + \exp\left(-\theta^\intercal x_i\right)}.
\]
It follows that 
\[
U_i(\theta) = -y_i\log h\left(\theta^\intercal x_i\right) - (1-y_i) \log h\left(-\theta^\intercal x_i\right).
\]

It is easy to see that
\[
\Abs{\partial_j U_i} = \Abs{(h(\theta^\intercal x_i)-y_i)x_{ij}} \le 1\cdot \Abs{x_{ij}}.
\] 
Thus we can set $M(\theta,\theta')$ to be $\norm{\theta-\theta'}_2$ and $c_i$ to be $\norm{x_{i}}_2$. We set the target acceptance rate to be $60\%$ and set $K$ to be around $\eps C/\max_ic_i$. The hyperparameters in baseline methods are set following their original papers~\citep{yildirim2019exact,heikkila2019differentially}.

\end{document}